%% file: main.tex
\documentclass[acmtog]{acmart}

\usepackage{booktabs} 
\usepackage{mathtools}
\usepackage{xfrac}
\usepackage{subcaption}
\usepackage[percent]{overpic}
\usepackage{xcolor}
\usepackage{multirow}

\usepackage{colortbl}
\usepackage{siunitx}

\usepackage[ruled]{algorithm2e} 

\SetAlFnt{\small}
\SetAlCapFnt{\small}
\SetAlCapNameFnt{\small}
\SetAlCapHSkip{0pt}

\input{macros.tex}
\setcopyright{acmlicensed}
\begin{document}
\title{\ourpapertitle}

\input{000_teaser_fig.tex}
\input{00_abstract}
%
%

\ccsdesc[500]{Computing methodologies~Computational photography}
\ccsdesc[500]{Computing methodologies~Image processing}
%
%

\acmJournal{TOG}
\acmYear{2019}\acmVolume{38}\acmNumber{6}\acmArticle{164}\acmMonth{11}
\acmDOI{10.1145/3355089.3356508}

\keywords{computational photography, low-light imaging}

\maketitle

\input{01_intro}
\input{03_motion_metering}
\input{04_motion_robustness}
\input{05_color}
\input{06_tone_mapping}
\input{07_results}
\input{08_conclusion}
\input{09_acknowledgement.tex}

\input{references.bbl}

\input{supplemental_text.tex}

\end{document}

%% file: macros.tex
\newcommand{\sect}[1]{Section~\ref{#1}}
\newcommand{\fig}[1]{Figure~\ref{#1}}
\newcommand{\tab}[1]{Table~\ref{#1}}
\newcommand{\eq}[1]{Equation~\ref{#1}}

\DeclarePairedDelimiter{\abs}{\lvert}{\rvert}
\DeclarePairedDelimiter{\norm}{\lVert}{\rVert}

\definecolor{darkgreen}{rgb}{0,0.5,0}
\definecolor{orange}{rgb}{1,0.5,0}
\definecolor{darkred}{rgb}{0.5,0,0}
\definecolor{Yellow}{rgb}{1,1, 0.6}
\definecolor{Orange}{rgb}{1,0.8, 0.6}
\definecolor{Red}{rgb}{1, 0.6, 0.6}

\newcommand{\ourpapertitle}{Handheld Mobile Photography in Very Low Light}

\author{Orly Liba}
\author{Kiran Murthy}
\author{Yun-Ta Tsai}
\author{Tim Brooks}
\author{Tianfan Xue}
\author{Nikhil Karnad}
\author{Qiurui He}
\author{Jonathan T. Barron}
\author{Dillon Sharlet}
\author{Ryan Geiss}
\author{Samuel W. Hasinoff}
\author{Yael Pritch}
\author{Marc Levoy}

\affiliation{
\institution{Google Research}
\streetaddress{1600 Amphitheatre Parkway}
\city{Mountain View}
\state{CA}
\postcode{94043}
}

%% file: 000_teaser_fig.tex
\begin{teaserfigure}
    \centering
    \includegraphics[width=\textwidth]{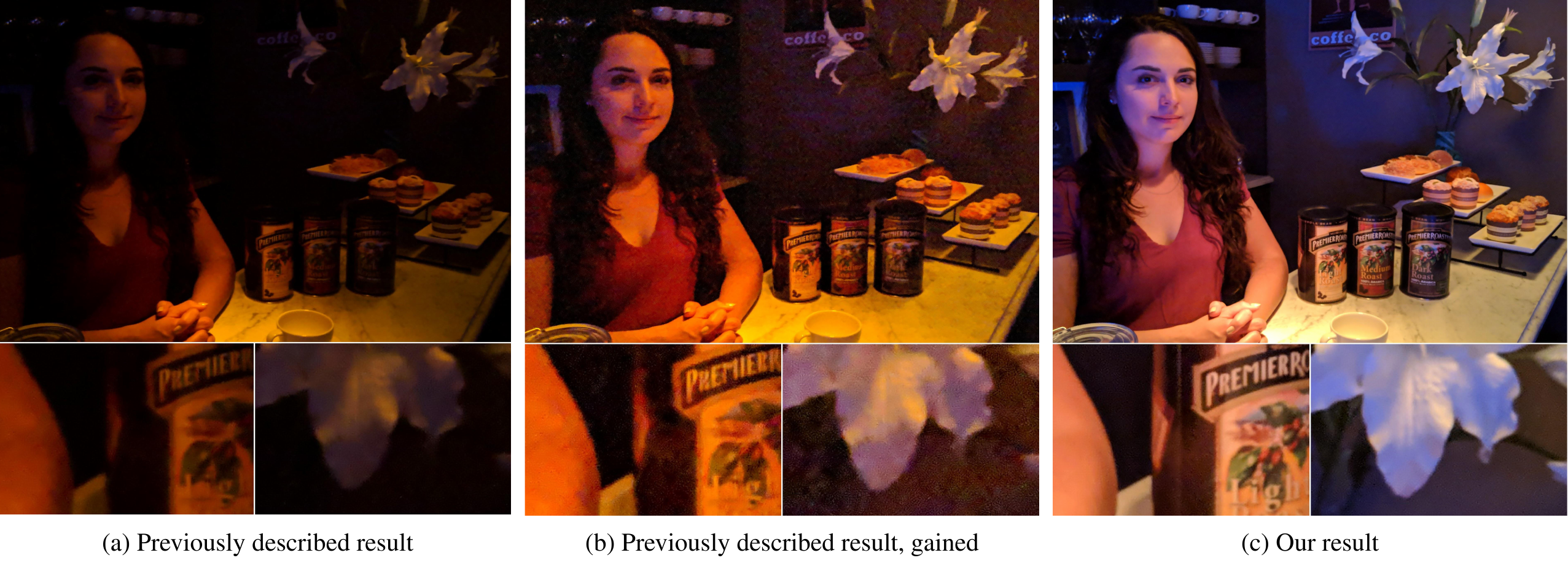}
    \caption{We present a system that uses a novel combination of motion-adaptive burst capture, robust temporal denoising, learning-based white balance, and tone mapping to create high quality photographs in low light on a handheld mobile device. Here we show a comparison of a photograph generated by the burst photography system described in \cite{hasinoff2016burst} and the system described in this paper, running on the same mobile camera. In this low-light setting (about 0.4 lux), the previous system generates an underexposed result (a). Brightening the image (b) reveals significant noise, especially chroma noise, which results in loss of detail and an unpleasantly blotchy appearance. Additionally, the colors of the face appear too orange. Our pipeline (c) produces detailed images by selecting a longer exposure time due to the low scene motion (in this setting, extended from $0.14$\,s to $0.33\,$s), robustly aligning and merging a larger number of frames (13 frames instead of 6), and reproducing colors reliably by training a model for predicting the white balance gains specifically in low light. Additionally, we apply local tone mapping that brightens the shadows without over-clipping highlights or sacrificing global contrast.}
    \label{fig:teaser}
\end{teaserfigure}

%% file: 00_abstract.tex
\begin{abstract}
Taking photographs in low light using a mobile phone is challenging and rarely produces pleasing results.
Aside from the physical limits imposed by read noise and photon shot noise, these cameras are typically handheld, have small apertures and sensors, use mass-produced analog electronics that cannot easily be cooled, and are commonly used to photograph subjects that move, like children and pets.
In this paper we describe a system for capturing clean, sharp, colorful photographs in light as low as 0.3~lux, where human vision becomes monochromatic and indistinct.
To permit handheld photography without flash illumination, we capture, align, and combine multiple frames.
Our system employs ``motion metering'', which uses an estimate of motion magnitudes (whether due to handshake or moving objects) to identify the number of frames and the per-frame exposure times that together minimize both noise and motion blur in a captured burst.
We combine these frames using robust alignment and merging techniques that are specialized for high-noise imagery.
To ensure accurate colors in such low light, we employ a learning-based auto white balancing algorithm.
To prevent the photographs from looking like they were shot in daylight, we use tone mapping techniques inspired by illusionistic painting: increasing contrast, crushing shadows to black, and surrounding the scene with darkness. All of these processes are performed using the limited computational resources of a mobile device.
Our system can be used by novice photographers to produce shareable pictures in a few seconds based on a single shutter press, even in environments so dim that humans cannot see clearly.
\end{abstract}

%% file: 01_intro.tex
\section{Introduction}
\label{sec:intro}

Low-light conditions present significant challenges for digital photography. This is a well-known problem with a long history of treatment in the camera and mobile device industries, as well as the academic literature.

The fundamental difficulty of low-light photography is that the signal measured by the sensor is low relative to the noise inherent in the measurement process. There are two primary sources of noise in these systems: shot noise and read noise \cite{nakamura2016image, macdonald2006digital}. If the signal level is very low, read noise dominates. As the signal level rises above the read noise, shot noise becomes the dominant factor. The signal levels relative to these noise sources, that is, the signal-to-noise ratio (SNR), needs to be managed carefully in order to successfully capture images in low light.

The most direct approaches for low-light photography have focused on increasing the SNR of the pixels read from the sensor. These strategies typically fall into one of these categories:

\begin{itemize}
    \item Expand the camera's optical aperture.
    \item Lengthen the exposure time.
    \item Add lighting, including shutter-synchronized flash.
    \item Apply denoising algorithms in post-processing.
\end{itemize}

Unfortunately, these techniques present significant hindrances for photographers---especially for mobile handheld photography, which is the dominant form of photography today. Increasing a camera's aperture increases the size of the device, making it heavier and more expensive to produce.  It also typically increases the length of the optical path, which conflicts with the market's preference for thin devices. Lengthening the exposure time leads to an increased chance of motion blur, especially when the device is handheld or if there are moving objects in the scene. Adding lighting makes it challenging to produce a natural look: the flash on mobile devices often makes the subject look over-exposed, while the background often looks black because it is too far away for the flash to illuminate. Post-processing single-frame spatial denoising techniques have been widely studied and are often very successful at improving the SNR of an image. However, when noise levels are very high, the reduction of noise often results in significant loss of detail. Moreover, merely improving the SNR addresses only some of the challenges of low light photography and does not yield an overall solution.

This paper presents a system for producing detailed, realistic-looking photographs on handheld mobile devices in light down to about 0.3~lux, a level at which most people would be reluctant to walk around without a flashlight.  Our system, which launched to the public in November 2018 as Night Sight on Google Pixel smartphones, is designed to satisfy the following requirements:

\begin{enumerate}
    \item Reliably produce high-quality images when the device is handheld, avoiding visual artifacts.
    \item Avoid using an LED or xenon flash.
    \item Allow the user to capture the scene without needing any manual controls.
    \item Support total scene acquisition times of up to ${\sim}6$ seconds.
    \item Accumulate as much light as possible in that time without introducing motion blur from handshake or moving objects.
    \item Minimize noise and maximize detail, texture, and sharpness.
    \item Render the image with realistic tones and white balance despite the mismatch between camera and human capabilities.
    \item Operate on a mobile device (with limited compute resources), with processing time limited to ${\sim}2$ seconds.
\end{enumerate}

While previous works have focused on specific aspects of low-light imaging (such as denoising \cite{dabov2007image,mildenhall2018burst}), there have been relatively few works that describe photographic \emph{systems} that address more than one of the above requirements. The main strategies that have been used by previous systems are burst imaging \cite{hasinoff2016burst}, and end-to-end trained convolutional neural networks (CNNs) \cite{chen2018learning}. Burst imaging systems capture, align, and merge multiple frames to generate a temporally-denoised image. This image is then further processed using a series of operators including additional spatial denoising, white balancing, and tone mapping. CNN-based systems attempt to perform as much of this pipeline as possible using a deep neural network with a single raw frame as its input. The network learns how to perform all of the image adjustment operators. CNN-based solutions often require significant computational resources (memory and time) and it is challenging to optimize their performance so that they could run on mobile devices. Additionally, end-to-end learning based systems don't provide a way to design and tune an imaging pipeline; they can only imitate the pipeline they were trained on.

Our solution builds upon the burst imaging system described in~\cite{hasinoff2016burst}. In contrast to CNN-based systems, \cite{hasinoff2016burst} was designed to perform well on mobile devices. Simply extending it to collect more frames allows it to generate high-quality photographs down to about 3~lux. However, to produce high-quality images in even lower light, tradeoffs between noise and motion blur must be addressed. Specifically, noise can be improved by increasing the exposure time, but individual frames might contain blur, and because the cited system does not perform deblurring, the motion blur present in the reference frame will remain in the output. Alternatively, noise can be improved by merging more frames and increasing the contribution of each merged frame, effectively increasing temporal denoising, however, doing so will cause moving objects to appear blurry in the output image.

\begin{figure*}
    \centering
    \includegraphics[width=\textwidth]{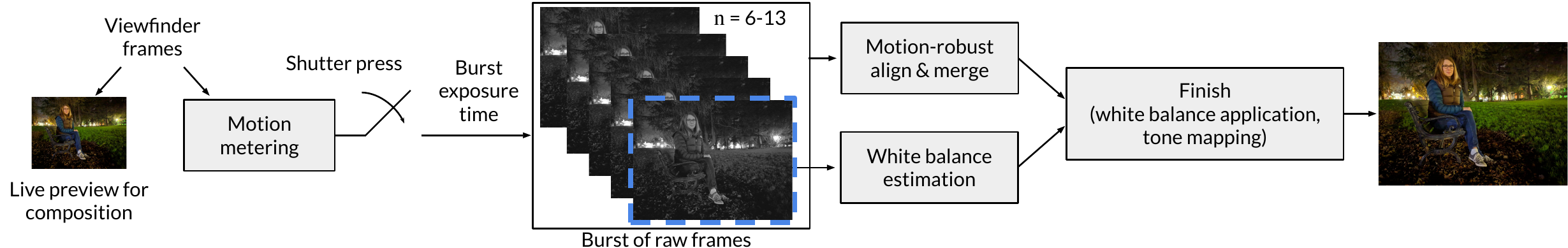}
    \caption{An overview of our processing pipeline, showing how we extend \cite{hasinoff2016burst}. Viewfinder frames are used for live preview for composition and for motion metering, which determines a per-frame exposure time that provides a good noise vs. motion blur tradeoff in the final result (Section~\ref{sec:motion_metering}). Based on this exposure time, we capture and merge a burst of 6--13 frames (Section~\ref{sec:motion_robustness}). The reference frame is down-sampled for the computation of the white-balance gains (Section~\ref{sec:awb}). White balance and tone mapping (Section~\ref{sec:tone_mapping}) are applied at the ``Finish'' stage, as well as demosaicing, spatial and chroma denoising and sharpening. Total capture time is between 1~and 6~seconds after the shutter press and processing time is under 2~seconds.}
    \label{fig:pipeline}
\end{figure*}

In addition to the problem of low SNR, low-light photographs suffer from a natural mismatch between the capabilities of a digital camera and those of the human visual system. Low light scenes often contain strongly colored light sources. The brain's visual processing can adapt to this and still pick out white objects in these conditions. However, this is challenging for digital cameras. In darker environments (such as outdoors under moonlight), a camera can take a clean, sharp, colorful pictures. However, we humans don't see well in these conditions: we stop seeing in color because the cone cells in our retinas do not function well in the dark, leaving only the rod cells that cannot distinguish between different wavelengths of light \cite{stockman2006into}. This scotopic vision system also has low spatial acuity, which is why things seem indistinct at night. Thus, a long-exposure photograph taken at night will look to us like daytime, an effect that most photographers (or at least most mobile camera users) would like to avoid.

The system described here solves these problems by extending~\cite{hasinoff2016burst} in the following ways:
\begin{enumerate}
    \item \textbf{Real-time motion assessment (``motion metering'')} to choose burst capture settings that best balance noise against motion blur in the final result. As an additional optimization, we set limits on exposure time based on an analysis of the camera's physical stability using on-device gyroscope measurements.
    \item \textbf{Improved motion robustness during frame merging} to alleviate motion blur during temporal denoising. Contributions from merged frames are increased in static regions of the scene and reduced in regions of potential motion blur. %
    \item \textbf{White balancing for low light} to provide accurate white balance in strongly-tinted lighting conditions. A learning-based color constancy algorithm is trained and evaluated using a novel metric.
    \item \textbf{Specialized tone mapping} designed to allow viewers of the photograph to see detail they could not have seen with their eyes, but to still know that the photograph conveys a dark scene.
\end{enumerate}

The collection of these new features, shown in the system diagram in \fig{fig:pipeline}, enables the production of low-light photographs that exceed human vision capabilities. The following sections cover these topics in detail. Section~\ref{sec:motion_metering} describes the motion metering system, and Section~\ref{sec:motion_robustness} describes the motion-robust temporal merging algorithm. Section~\ref{sec:awb} details how we adapt a learning-based color constancy algorithm to perform white balancing in low-light scenes. Section~\ref{sec:tone_mapping} presents our tone-mapping strategy. Section~\ref{sec:results} and Section~\ref{sec:results_appendix} of the appendix show the results of our system for different scenes and acquisition conditions as well as comparisons to other systems and devices. All images were captured on commercial mobile devices unless otherwise specified. Most of the images in the paper were captured while handheld. If a photo was captured using a tripod or braced, it is indicated as "camera stabilized" in the caption.

To further the exploration of low-light photography, we have released a datatset of several hundred raw bursts of a variety of scenes in different conditions captured and processed with our pipeline \cite{NSWebsite}.

%% file: 03_motion_metering.tex
\section{Motion metering}
\label{sec:motion_metering}

\begin{figure*}
    \centering
    \includegraphics[width=0.9\textwidth]{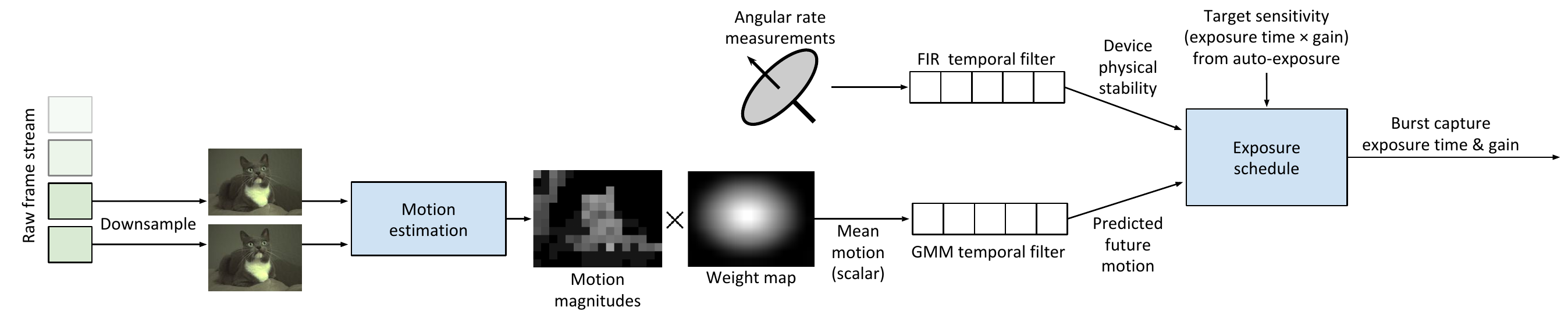}
    \caption{Overview of the motion metering subsystem. The goal is to select an exposure time and gain (ISO) that provides the best SNR vs. motion blur tradeoff for a burst of frames captured after the shutter press. This is done by measuring scene and camera motion before the shutter press, predicting future motion, and then using the prediction to calculate the capture settings. To measure motion, a stream of frames is downsampled and then processed by an algorithm that directly estimates the magnitude of the optical flow field (``Bounded Flow''). A weighted average of the flow magnitude is then calculated, assigning higher weights to areas where limiting motion blur is more important. To predict future motion, the sequence of weighted averages is processed by a temporal filter based on a Gaussian mixture model (GMM). Concurrently, angular rate measurements from a gyro sensor are filtered to assess the device's physical stability. These measurements are used to build a custom ``exposure schedule'' that factorizes the sensor's target sensitivity computed by a separate auto-exposure algorithm into exposure time and gain. The schedule embodies the chosen tradeoff between SNR and motion blur based on the motion measurements.}
    \label{fig:motion_metering_system_diagram}
\end{figure*}

To capture a burst of images on fixed aperture mobile cameras, the exposure time and gain (ISO) must be selected for each frame in the burst. Here, the gain represents the combination of analog and digital gain. However, several constraints limit the degrees of freedom within these parameters:

\begin{itemize}
    \item We leverage the strategy in \cite{hasinoff2016burst}, where all frames in the burst are captured with the same exposure time and gain after the shutter press, making it easier to align and merge the burst.
    \item An auto-exposure system (e.g., \cite{schulz2007using}) determines the sensor's target sensitivity (exposure time $\times$ gain) according to the scene brightness.
    \item As mentioned in Section~\ref{sec:intro}, the total capture time should be limited ($\leq$ 6~seconds for our system).
\end{itemize} 

Thus, selecting the capture settings reduces to decomposing the target sensitivity into the exposure time and gain, and then calculating the number of frames according to the capture time limit (i.e.\ 6~seconds divided by the exposure time, limited to a maximum number of frames set by the device's memory constraints). \cite{hasinoff2016burst} performs the decomposition using a fixed ``exposure schedule'', which conservatively keeps the exposure time low to limit motion blur. However, this may not yield the best tradeoff in all scenarios.

In this section, we present ``motion metering'', which selects the exposure time and gain based on a prediction of future scene and camera motion. The prediction is used to shape an exposure schedule dynamically, selecting slower exposures for still scenes and faster exposures for scenes with motion. The prediction is driven by measurements from a new motion estimation algorithm that directly computes the magnitude of the optical flow field. Only the magnitude is needed since the direction of motion blur has no effect on final image quality.

A key challenge is measuring motion efficiently and accurately, as the motion profile in the scene can change rapidly. This can occur due to hand shake or sudden changes in the motion of subjects. Recent work in optical flow has produced highly accurate results using convolutional neural networks (\cite{EpicFlow,FlowNet}), but these techniques have a high computational cost that prohibits real-time performance on mobile devices. Earlier optical flow algorithms such as \cite{Lucas81} are faster but less accurate.

Another key challenge is how to leverage motion measurements to select an exposure time and gain. \cite{portz2011} optimizes the exposure time to limit motion blur exclusively at the expense of noise, and \cite{gyrad2015} uses a heuristic decision tree driven by the area taken up by motion in the scene. \cite{boracchi2012} provides a technique to optimize the exposure time given a camera motion trajectory and known deblurring reconstruction error in different situations. However, this is primarily useful when strong deblurring can be applied without producing artifacts (a challenging open problem).

\fig{fig:motion_metering_system_diagram} depicts our subsystem that addresses these challenges. It combines efficient and accurate motion estimation, future motion prediction, and a dynamic exposure schedule that allows us to tune the system to prioritize noise vs. motion blur. Flow magnitudes are estimated between $N$ successive pairs of frames before shutter press (\sect{sec:motion_estimation}). Each flow magnitude solution is center-weighted averaged, and the sequence of averages is used to fit a probabilistic model of the motion profile. The model is then used to calculate an upper bound of the minimum motion over the next $K$ frames, where $K$ is chosen such that the motion blur of the merged burst is below some threshold (\sect{sec:motion_prediction}). Finally, this bound is used to shape the exposure schedule (\sect{sec:exposure_time_selection}). We also utilize angular rate estimates from gyro measurements to enhance the future motion prediction (\sect{sec:tripod_detection}), which improves the robustness of our system in very low light conditions.

\subsection{Efficient motion magnitude estimation}
\label{sec:motion_estimation}

We present a new algorithm that directly computes the magnitude of the optical flow field. We start with the Taylor expansion of the brightness constancy constraint \cite{negahdaripour1993generalized}:
\begin{equation}
    \Delta I_t(x, y) = \vec{g}(x, y)^\mathrm{T} \vec{v}(x, y)\ ,
\end{equation}
the change in intensity between a pair of images at the same pixel location $\Delta I_t$ is equal to the inner product of the spatial intensity gradient at that location $\vec{g}$ and the motion vector at that location $\vec{v}$. The Cauchy-Schwartz inequality can be applied to this expression, and then rearranged to bound the magnitude of the motion vector:
\begin{equation}
\begin{aligned}
    |\Delta I_t(x, y)| &\leq \norm{\vec{g}(x, y)} \cdot \norm{\vec{v}(x, y)} \\
    \norm{\vec{v}(x, y)} &\geq \frac{|\Delta I_t(x, y)|}{\norm{\vec{g}(x, y)}}
    \label{eq:bounded_flow}
\end{aligned}
\end{equation}
Thus, a lower bound for the flow magnitude can be computed by dividing the magnitude of the change in intensity by the norm of the spatial gradient. We call this technique ``Bounded Flow". The bound reaches equivalence when the direction of motion is parallel to the gradient. This is the same property as the aperture problem in full flow solutions: flow along edges can be under-estimated if the direction of motion has a component perpendicular to the gradient. Thus, the ``lower bound'' in \eq{eq:bounded_flow} can be considered a direct computation of the optical flow magnitude according to Lucas-Kanade.

\begin{figure}
    \centering
    \begin{subfigure}[c]{0.34\linewidth}
        \includegraphics[width=\linewidth]{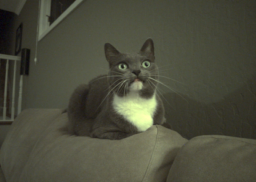}
        \caption{Input}
        \label{subfig:mm_image_pair}
    \end{subfigure}
    ~
    \begin{subfigure}[c]{0.32\linewidth}
        \includegraphics[width=\linewidth]{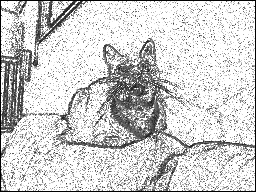}
        \caption{Motion magnitude}
        \label{subfig:mm_unmasked}
    \end{subfigure}
    ~
    \begin{subfigure}[c]{0.32\linewidth}
        \includegraphics[width=\linewidth]{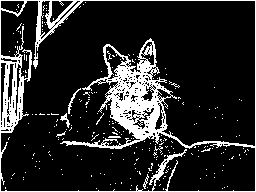}
        \caption{Noise mask}
        \label{subfig:mm_mask}
    \end{subfigure}
    
    \begin{subfigure}[t]{0.48\linewidth}
        \includegraphics[width=\linewidth]{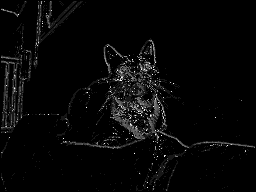}
        \caption{Masked motion magnitude}
        \label{subfig:mm_after_mask}
    \end{subfigure}
    ~
    \begin{subfigure}[t]{0.48\linewidth}
        \includegraphics[width=\linewidth]{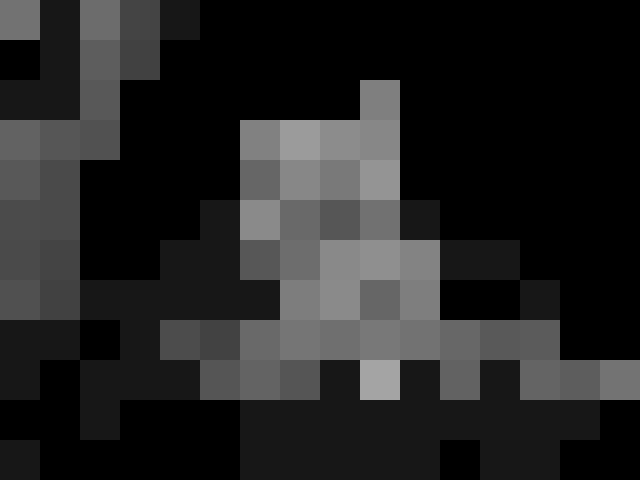}
        \caption{Refined motion magnitude}
        \label{subfig:mm_final}
    \end{subfigure}
    \caption{Intermediate results of ``motion metering''. The input is a pair of downsampled linear images with a known noise variance (\subref{subfig:mm_image_pair}). The white/black levels and the relative exposures are used to normalize the images so that \eq{eq:bounded_flow} can be applied. The result is (\subref{subfig:mm_unmasked}), which shows the motion magnitude in a grayscale image (white = 10~pixels of motion). The motion is overestimated in areas with low gradients. The noise model is used to compute areas where the gradient is high relative to the noise, and this serves as a mask (\subref{subfig:mm_mask}) where valid motion measurements can be made (\eq{eqn:noise_mask}). The motion magnitudes after masking are shown in (\subref{subfig:mm_after_mask}). These are further refined by downsampling with outlier rejection, taking the 90th-percentile motion in each bin. The final result is shown in (\subref{subfig:mm_final}). The resolution of the final output is 16$\times$12.}
    \label{fig:noise_mask}
\end{figure}

The flow magnitude is computed using a pair of linear (non-tone mapped) images on a per-pixel basis. The linearity of the signal enables the efficient modeling of pixel noise variance, where the variance is a linear function of the signal level \cite{nakamura2016image}. We leverage this by masking out motion in regions where the gradient $\norm{\vec{g}}$ is too low relative to the noise standard deviation $\sigma$:
\begin{equation}
    \norm{\vec{g}(x, y)} < K\sigma\ .
    \label{eqn:noise_mask}
\end{equation}
We use $K=2.5$. Discarding small spatial gradients is particularly important in low-light scenes, where noise can be prevalent and can corrupt the flow magnitude estimate. As a final step, the motion magnitudes are refined by downsampling and taking the 90-th percentile motion in each bin, as a form of outlier rejection. \fig{fig:noise_mask} shows the effect of applying the noise mask and the refinement.

Table \ref{tab:metering_results} shows the accuracy of the motion magnitudes computed by Bounded Flow relative to other methods, computed from the MPI-Sintel optical flow dataset \cite{Butler2012}. The methods' run times are also provided. Bounded Flow with noise masking compares favorably to other flow methods when comparing motion magnitudes estimated from noisy imagery.  Bounded Flow is more accurate than other methods that do not use CNNs, but is also faster by at least 2$\times$ over Lucas-Kanade based methods.

\begin{table}
\caption{Comparison of different optical flow algorithms. The means of absolute errors (MAEs) between the predicted and true motion magnitudes were generated using the MPI-Sintel optical flow dataset \cite{Butler2012}. The images were inverse-tone mapped into a simulated linear space using \cite{Unprocessing}, and high-magnitude noise was added, with a standard deviation of 5\% of the white level. The images were then downsampled to 256$\times$192 using a triangle kernel before being processed by the algorithms. Errors are computed only at locations indicated by the noise-based mask in \eq{eqn:noise_mask}. Run times are also provided, measured from single-threaded executions running on an Intel Xeon E5-2690 v4 CPU. We see that Bounded Flow is as accurate as other methods that are not based on neural networks. Furthermore, Bounded Flow is 2$\times$ faster than the next fastest method. This makes it suitable for our real-time application where the motion magnitude needs to be monitored constantly on a mobile device while drawing minimal power.
}
\begin{tabular}{l| S[table-format=3.2] | S[table-format=3.2] }
    & \multicolumn{1}{c |}{MAE} & \multicolumn{1}{c}{Run time} \\
    Motion estimation method & \multicolumn{1}{c |}{(pixels)} & \multicolumn{1}{c}{(ms)} \\
    \hline \hline 
    Coarse-to-fine flow \cite{liu2009beyond} & 0.74 & 310.0 \\
    FlowNet \cite{FlowNet} & \phantom{\cellcolor{Yellow}} 0.55 & 451.0 \\
    Non-iterative Lucas-Kanade (LK) & 0.89 & 6.25 \\
    Iterative LK (max 20 iter) & 0.73 & 9.45 \\
    Iterative pyramid LK (max 20 iter, 3 lvls) & 0.81 & 17.0 \\
    \hline
    Our Bounded Flow algorithm & 0.62 & \phantom{\cellcolor{Yellow}} 3.11
\end{tabular}

\label{tab:metering_results}
\end{table}

After motion magnitude has been measured across the scene, the measurements are aggregated into a single scalar value by taking a weighted average. Typically, a center-weighted average is used, where motion in the center of the scene is weighted highly, and motion on the outside of the scene is discounted. Humans viewing photographs are particularly perceptive to motion blur on faces, therefore, if a face is detected in the scene by a separate automatic system such as \cite{howard2017mobilenets}, then the area of the face is weighted higher compared to the rest of the scene. The same type of weighting occurs if a user manually selects an area of the scene. Spatial weight maps of this kind are common in many auto-exposure and autofocus systems.

\subsection{Future motion prediction}
\label{sec:motion_prediction}

Given a sequence of weighted average motion measurements, we predict the scene motion of future frames comprising the post-shutter burst. The predictor considers a key property of temporal denoising systems such as~\cite{hasinoff2016burst}: one frame out of the first $K$ frames of the burst is selected as the ``reference frame'', and is denoised by merging into it other frames in the burst. The end result is that the motion blur of the merged frame is close to the motion blur of the reference frame, and thus, the frame with the least motion blur should be selected as the reference frame. Thus, we can pose the future motion prediction problem as the following: given motion samples $v_i,~i=1,2,\cdots,N$ before the shutter press, what is the minimum motion $v_{\text{min}}$ within the next $K$ frames.

One way to predict future motion is to use a finite impulse response (FIR) filter on the past $N$ samples. There are two issues with this approach. First, FIR filters are not very robust to outliers. If one of the past frames has extremely large motion, either due to a fast moving object or inaccurate motion estimation, the FIR filter output will suddenly increase. However, the outlier frame can easily be rejected when selecting the reference frame, so this should not suddenly lower the final selected exposure time. Second, FIR filters do not consider that only one out of $K$ frames are used as a reference and they require tuning when $K$ changes. For example, when $K$ increases, the maximum exposure time should increase, since there is a higher chance that one frame out of the larger pool of frames remains sharp. However, re-tuning an FIR filter to exhibit this behavior is not straightforward.

To resolve these issues, we propose a motion prediction method based on a cumulative distribution function (CDF). We find the minimum motion $v_{\text{min}}$ such that:
\begin{equation}
Pr\left[v_{\text{min}} \geq \min_{k=1}^{K} v_k \,\,\Big|\,\, \{v_i\}_{i=1,\cdots,N}\right] \geq P_{\text{conf}}\ .
\label{eqn:motion_prediction}
\end{equation}
That is, $v_{\text{min}}$ is an upper bound on the minimum motion over the next $K$ frames, to a high degree of confidence ($P_{\text{conf}}$). An upper bound was chosen in order to obtain a conservative estimate of how much the exposure time can be lengthened, thus reducing the chance of unwanted motion blur.

To model the probability of scene motion after the shutter press, we assume that future motion shares the same distribution as scene motion before shutter press. We fit a three-cluster Gaussian mixture model (GMM) to the weighted average motion samples before shutter press using expectation-maximization \cite{Dempster1977MaximumLF}. Then, assuming that each motion magnitude $v_k$ is independently drawn from the GMM, we have:
\begin{equation}
1 - Pr\left[v_{\text{min}} \leq v_k\right]^K \geq P_{\text{conf}}\ . \label{eqn:motion_prediction2}
\end{equation}
Since $v$ is a 1D GMM, we can calculate $Pr[v_{\mathrm{min}} \leq v_k]$ for each Gaussian cluster by using the unit normal table. To find the minimum motion $v_{\mathrm{min}}$ such that \eq{eqn:motion_prediction2} still holds, we use binary search because the left hand side of \eq{eqn:motion_prediction2} is a monotonic function with respect to $v_{\mathrm{min}}$. \tab{tab:motion_filters} shows that this GMM-based filter is more accurate at future motion prediction than FIR filtering.

\begin{table}
\caption{Comparison of three different filters for future motion prediction, evaluated on a dataset of 777 sequences of weighted average motion magnitude estimates from Bounded Flow taken from sequential frames when composing real-world scenes. For each sequence, we use the motion estimates between $N$ successive frames to predict the minimum pairwise motion of next $K$ frames (here, $N=10$ and $K=4$). We compare:  1) the averaging of 5 motion samples, 2) a 8-tap Hamming-window FIR filter with a cutoff frequency of 0.01 of Nyquist, and 3) our Gaussian mixture model (GMM)-based probabilistic prediction on two evaluation metrics: RMSE (root-mean-square error) and RMSE on log2 motion. We calculate log2 to roughly denote f-stops, as the ideal exposure time is inversely proportional to motion magnitude \eq{eqn:blur_limiting_texp}. On both metrics, the GMM filter has the lowest error among all three algorithms.}

\begin{tabular}{c|cc|c}
    & mean filter & FIR filter & GMM \\ \hline \hline
    RMSE & 1.2652 & 1.4671 & \phantom{\cellcolor{Yellow}} 1.1722 \\
    RMSE of log2 motion & 0.2538 & 0.2967 & \phantom{\cellcolor{Yellow}} 0.1963
\end{tabular}

\label{tab:motion_filters}
\end{table}

\subsection{Camera stability detection}
\label{sec:tripod_detection}

\begin{figure}
\centering
\includegraphics[width=0.98\linewidth]{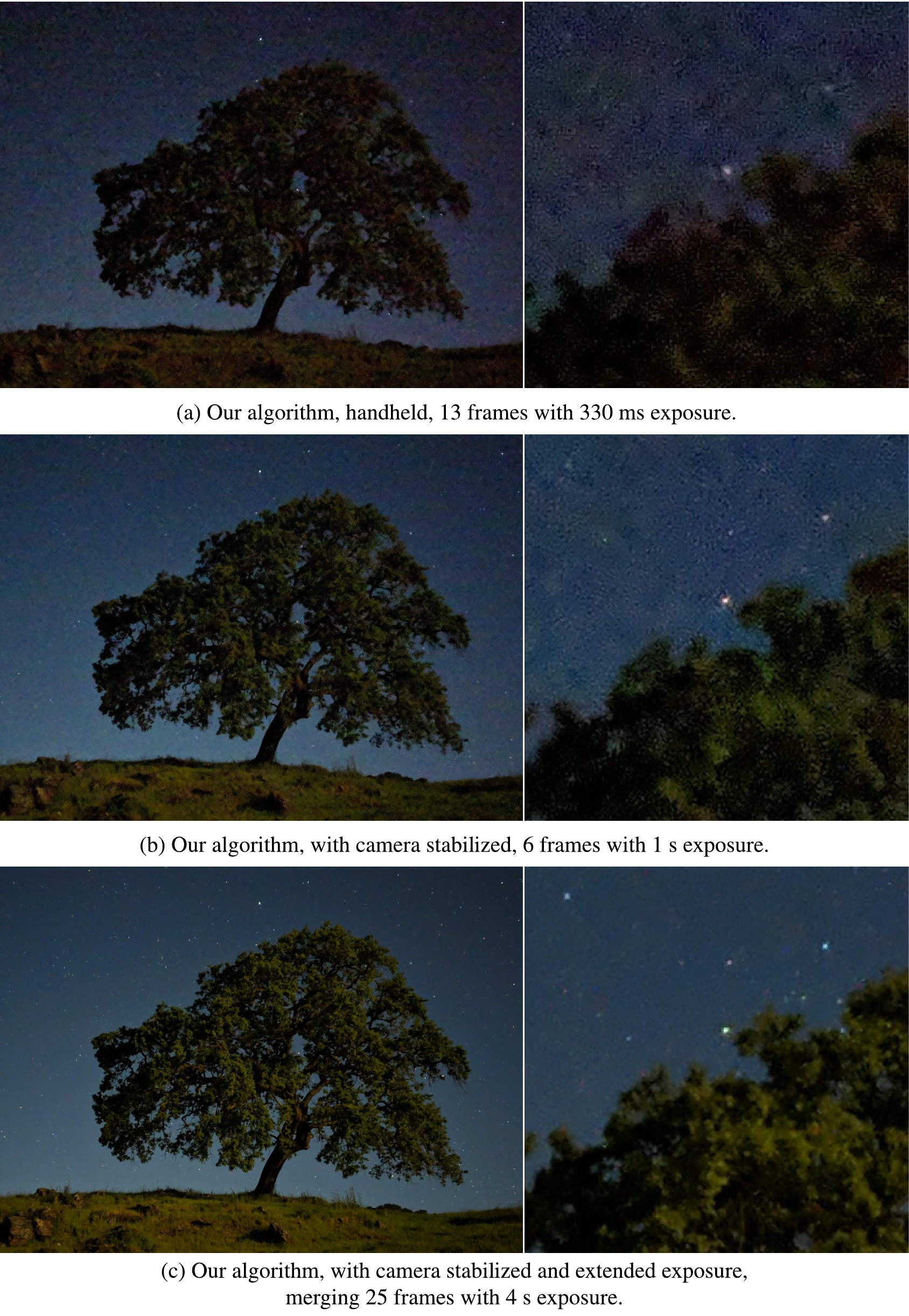}
\caption{A demonstration of our algorithm with extended exposures enabled by camera stability detection. The measured light levels at the tree were 0.06~lux. (a) A handheld capture uses $13\times$ $333\,\mathrm{ms}$ exposures. (b) When stabilized on a tripod, our algorithm uses $6\times$ $1\,\mathrm{s}$ exposures. The longer individual exposures greatly reduce noise. (c) Camera stability detection could be used for even longer captures, such as the $25\times$ $4\,\mathrm{s}$ exposures in this example. Noise is significantly reduced, at the cost of a much longer total capture time.
}
\label{fig:images_with_extended_exposure}
\end{figure}

\begin{figure}
\begin{subfigure}[b]{.5\linewidth}
\centering \includegraphics[width=0.97\linewidth]{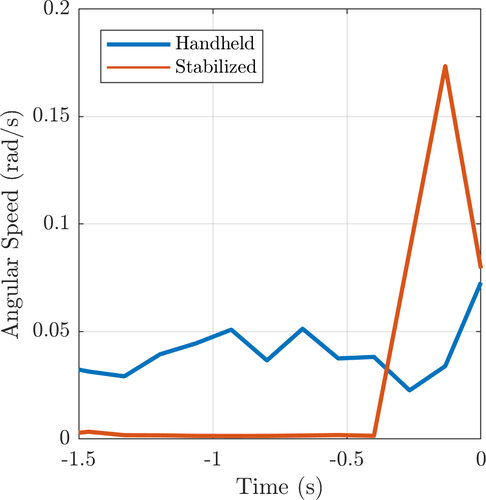}
\caption{Angular speed, raw signal}
\end{subfigure}%
\begin{subfigure}[b]{.5\linewidth}
\centering \includegraphics[width=0.97\linewidth]{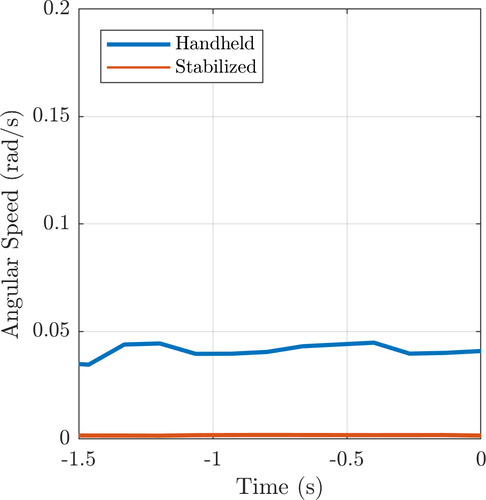}
\caption{Angular speed, filtered}
\end{subfigure}%
\caption{Masking and smoothing of a gyroscope signal for a more reliable camera stability detection. (a) shows the raw gyroscope signals leading up to the capture of handheld and stabilized bursts taken of the same scene. Despite the camera being stabilized on a tripod, angular speed peaks right before capture due to the shutter press. As described in Section \ref{sec:tripod_detection}, motion caused by the shutter press can briefly meet or exceed handheld motion, making reliable camera stability detection challenging. (b) shows the temporally smoothed and masked gyroscope signals, to ignore the brief motion from shutter press. This results in more accurate and reliable discrimination between handheld and stabilized cameras.}
\label{fig:camera-motion}
\end{figure}

In low light, experienced photographers will often stabilize their camera on a tripod or against a surface to allow for longer exposure times. While we prioritize handheld capabilities, we also allow longer exposures if a user chooses to brace the device, leave it propped, or put the device on a tripod. When the device is handheld, we capture up to $333$~ms exposures; when we detect the device is stabilized, we capture up to $1$~s exposures. \fig{fig:images_with_extended_exposure} shows a very dark scene captured using a tripod at different exposure times. %

Before extending the exposure time, we must be certain that the camera is braced or on a tripod. To accomplish this, we leverage measurements from the device's gyroscope sensor. Gyroscopes provide a useful signal in video stabilization \cite{videostabilization, karpenko2011digital} and alignment \cite{ringaby14virtualtripod}, and they are similarly valuable for detecting camera stability. Moreover, they provide reliable measurements even in very low light, in which longer exposures have the most benefit.

To assess camera stability from gyro measurements, we temporally smooth angular speeds by averaging them over the $t_0=1.466$\,s before shutter press. When on a tripod, significant vibration can be caused by the shutter button press, which, on our device, we determined empirically to last for a few hundred milliseconds. This vibration is particularly large if the camera is on a small, lightweight tripod, and can even exceed the camera motion when handheld. Since there is some latency between the physical shutter press and the software receiving this signal, some of these measurements are counted as pre-shutter angular speed. The combination of these factors makes it difficult to reliably differentiate between handheld and stabilized cameras. We therefore mask out the $t_1=0.400$\,s of gyroscope readings nearest to the shutter press signal.

The camera motion measurement is thus the average angular speed in the range $-t_0 \leq t \leq -t_1$, where $t=0$ represents the time at which capture begins. Figure~\ref{fig:camera-motion} shows the improved reliability of gyroscope measurements after temporal smoothing and shutter press masking. Any residual motion in the first frame of the burst caused by the shutter press will be rejected by reference frame selection.

\subsection{Exposure time selection}
\label{sec:exposure_time_selection}

Based on motion prediction and camera stability detection, our goal is to select the exposure time and gain that will best trade off SNR for motion blur in the final rendered image. This becomes more challenging in low-light scenes with fast-moving subjects. Selecting a short exposure time that freezes the motion would cause excessive noise, and selecting long exposure times would cause too much motion blur. Therefore, in these scenarios, a balance needs to be struck between the two extremes.

\begin{figure}
    \centering
    \includegraphics[width=\linewidth]{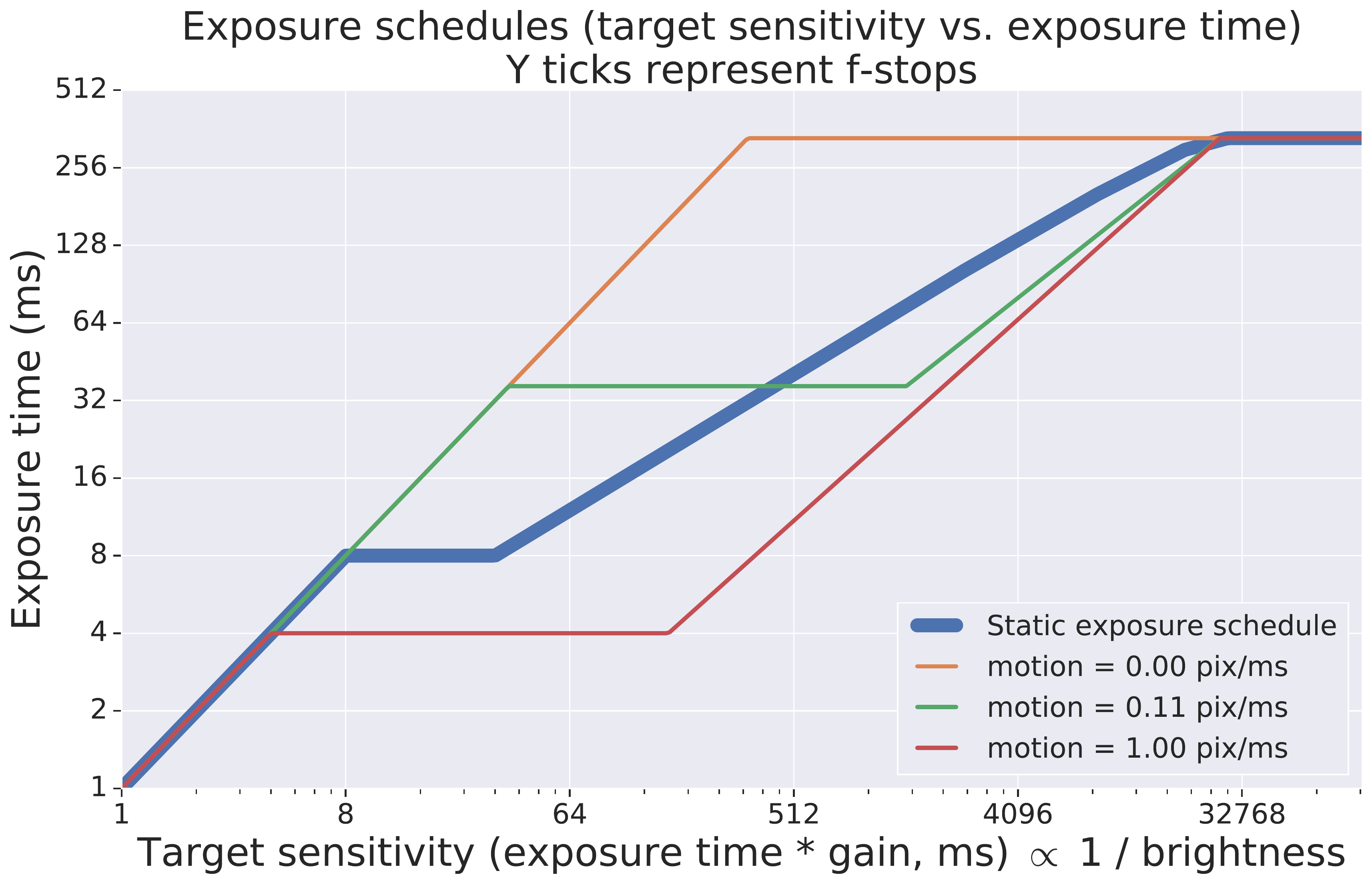}
    \caption{Cameras typically encode the exposure time vs. gain tradeoff with a static ``exposure schedule'', depicted by the thicker blue line. If the exposure schedule curve is higher, the camera favors higher SNR and more motion blur, and if the curve is lower, the camera favors lower SNR and less motion blur. Instead of a static schedule, we use a dynamic schedule based on the measured scene motion. This allows the schedule to provide a better tradeoff for the scene's conditions. Zero motion (orange line) leads to the slowest possible schedule, and as motion increases, the schedule favors proportionally faster exposure times. The flat regions of the dynamic exposure schedules denote the ``blur-limiting exposure time" in \eq{eqn:blur_limiting_texp}.}
    \label{fig:exposure_schedule}
\end{figure}

To address this, we use an exposure schedule that factorizes the target sensitivity (generated by an auto-exposure algorithm) into exposure time and gain. \fig{fig:exposure_schedule} depicts a typical exposure schedule for a mobile camera. Instead of using a static exposure schedule, we use a dynamic schedule that is shaped according to the scene motion and camera stability estimates, shown in \fig{fig:exposure_schedule}. The camera stability estimate governs the maximum exposure time returned by the schedule. Then, the motion prediction is used to compute a ``blur-limiting exposure time'', which restricts the estimated motion blur to $B$ pixels:
\begin{equation}
    t_{\mathrm{exp}}^{(B)} = \frac{B}{v_{\mathrm{min}}}\ ,
    \label{eqn:blur_limiting_texp}
\end{equation}
where $v_{\mathrm{min}}$ is calculated from Equation~\ref{eqn:motion_prediction2}. As the scene becomes darker, higher gains are required to restrict the exposure time to this value. At a certain point, the gain starts to result in excessive noise, and thus, the exposure schedule begins to increase both exposure time and gain together. Although this can cause the captured images to exceed the motion blur threshold, the noise remains reasonable in the final result.

The difference between static and dynamic schedules depends on the light level. In bright scenes ($> 100$~lux), the SNR can be high even with short exposure times, so the gain is kept at the minimum value in both schedules. In very dark scenes ($< 0.3$~lux), both schedules choose the maximum exposure time. The lux levels in between are where the dynamic schedule helps the most.

\begin{figure}
    \centering
    \includegraphics[width=\linewidth]{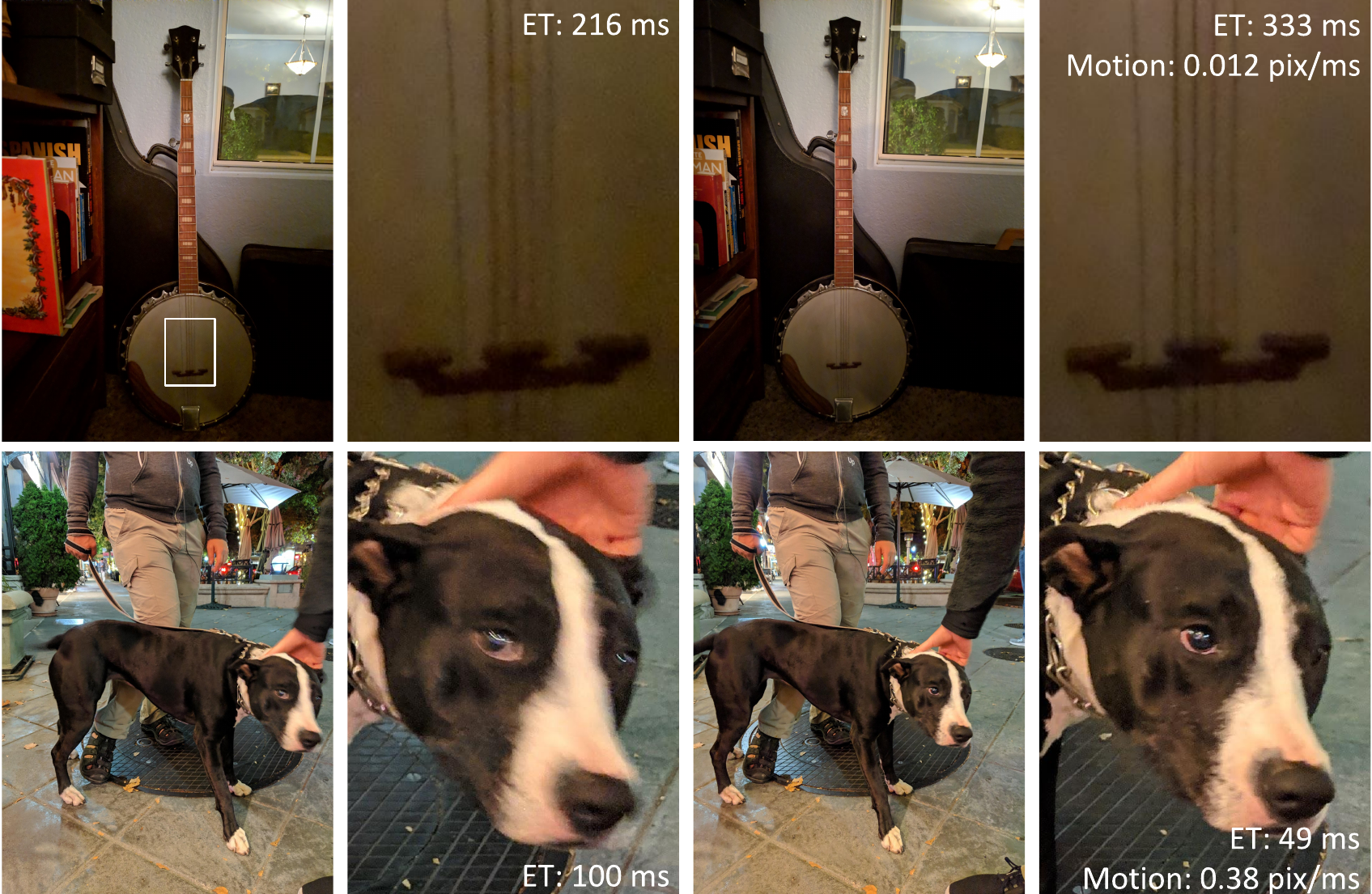}
    \caption{Comparison of images captured with static (left) and dynamic (right) exposure schedules. The pairs were captured at the same time with different exposure times governed by the schedules. The top row shows the comparison on a still scene (although there is some hand motion) where the dynamic schedule selects an exposure time $0.5$~f/stop longer than the static schedule, thereby reducing noise and increasing detail without incurring additional motion blur. The bottom row shows a comparison on a scene with significant subject motion, for which the dynamic schedule selects an exposure time $1$~f/stop shorter, thereby reducing motion blur without adding noticeable noise.}
    \label{fig:exposure_schedule_images}
\end{figure}

\fig{fig:exposure_schedule_images} shows images collected at the same time with static and dynamic exposure schedules. They indicate that the dynamic schedule chooses better points in the SNR vs. motion blur tradeoff than the static schedule.

%% file: 04_motion_robustness.tex
\section{Motion-adaptive burst merging}
\label{sec:motion_robustness}

Aligning and merging a burst of frames is an effective method for reducing noise in the final output. A relatively simple method that aligns and averages frames was implemented in SynthCam \cite{synthcam}. While this method is very effective at reducing noise, moving subjects or alignment errors will cause the resulting image to have motion artifacts. Convolutional neural networks \cite{mildenhall2018burst, Godard_2018, aittala2018burst} are a more recent and effective method for motion-robust frame merging. However, these techniques require considerable computation time and memory, and are therefore not suited to running within 2~seconds on a mobile device. Bursts can be aligned and merged for other applications, such as super-resolution~\cite{IRANI1991231, milanfarsuperres2004, sabre2019}. These techniques can also be beneficial for denoising in low light and we have provided additional information on using \cite{sabre2019} in Section~\ref{sec:merge_SR} of the appendix.

Since our method must be fast and robust to motion artifacts, we build upon the Fourier domain temporal merging technique described in \cite{hasinoff2016burst}, which was targeted for mobile devices. Other fast burst denoising techniques include \cite{liu2014burstdenoising}, \cite{delbracio2015video}, \cite{dabov2007image}, and \cite{maggioni2012bm4d}. However, these methods operate on tone-mapped (JPEG) images instead of raw images. As a result, \cite{hasinoff2016burst} benefits from increased dynamic range and better adaptation to different SNR levels in the scene due to having a more accurate noise variance model. In addition, \cite{liu2014burstdenoising} performs per-pixel median selection instead of denoising a reference frame, which can lead to artifacts at motion boundaries and locations where different objects with similar intensities occur. For an in-depth comparison of these fast burst denoising techniques, the reader is referred to the supplementary material from \cite{hasinoff2016burst}.

Fourier domain merging has a fundamental trade off between denoising strength and motion artifacts. That is, if the scene contains motion, it is challenging to align different frames correctly, and a simple average of the aligned frames would result in motion blur and ghosting. \cite{hasinoff2016burst} addresses this by reducing the degree of merging (i.e., reducing weights in a weighted average) from frequency bands whose difference from the reference frame cannot be explained by noise variance. However, the reduction of the weights is controlled by a global tuning parameter, and as a result, must be tuned conservatively to avoid motion artifacts. Consequently, more denoising could occur in static regions of scenes without incurring artifacts.

In low-light scenes, it becomes important to recover the lost denoising performance in static regions of scenes while remaining robust to motion artifacts. To do so, we propose adding a spatial domain similarity calculation (in the form of ``mismatch maps'', defined below) before frequency domain merging. The mismatch maps allow us to specify where the frequency-based merging can be more or less aggressive, on a per-tile basis. Using this method we can increase temporal merging in static regions in order to reduce noise, while avoiding artifacts by reducing merging in dynamic regions and in tiles which may have been aligned incorrectly.

Our contributions on top of \cite{hasinoff2016burst} are 1) Calculating mismatch maps, in which high mismatch represents dynamic regions in the scene; 2) Using these maps to enable spatially varying Fourier-domain burst merging, in which merging is increased in static regions and reduced in dynamic regions; and 3) Increasing spatial denoising where merging was limited, to compensate for the reduced temporal denoising.

\subsection{Alignment}

Alignment and merging are performed with respect to a reference frame, chosen as the sharpest frame in the burst. We use a tile-based alignment algorithm similar to \cite{hasinoff2016burst} with slight modifications to improve performance in low light and high noise. The alignment is performed by a coarse-to-fine algorithm on four-level Gaussian pyramids of the raw input. In our implementation, the size of the tiles for alignment is increased as the noise level in the image rises, at the cost of slightly increased computation for larger tile sizes. Whereas the previous alignment method used $8$ or $16$ pixel square tiles, we use square tiles of $16$, $32$, and $64$ pixels, depending on the noise level. For example, in a well-lit indoor setting, the size of the alignment tiles is $16$ while in the dark it is $64$ pixels.

\subsection{Spatially varying temporal merging}

As described in \cite{hasinoff2016burst}, after alignment, tiles of size $16 \times 16$ pixels are merged by performing a weighted average in the Fourier domain. This merging algorithm is intrinsically resilient to temporal changes in the scene and to alignment errors, owing to the reduced weight given to frames and frequency bands that differ from the reference frame significantly above the noise variance. In \cite{hasinoff2016burst}, the weight of the contribution from each frame $z$ in the location of tile $t$, is proportional to $1 - A_{tz}(\omega)$, in which $A_{tz}(\omega)$ is
\begin{align}
    A_{tz}(\omega) &= \frac{\abs{D_tz(\omega)}^2}{\abs{D_tz(\omega)}^2 + c{\sigma_{tz}}^2}\ ,  \label{eq:HDR+merge}
\end{align}
where $D_tz(\omega)$ is the difference between tiles in Fourier band $\omega$ as defined in \cite{hasinoff2016burst}, ${\sigma_{tz}}^2$ is the noise variance provided by a model of the noise, and $c$ is a constant that accounts for the scaling of the noise variance. If the difference between tiles is small relative to the noise, $A_{tz}(\omega)$ decreases and a lot of merging occurs, and if the difference is large, the function goes to 1 and the contribution of the tile decreases.

In \eq{eq:HDR+merge}, the value of $c$ can be tuned to control the tradeoff between denoising strength and robustness to motion and alignment-error artifacts. We call this parameter the ``temporal strength'', where a strength of 0 represents no temporal denoising, and a strength of infinity represents averaging without any motion robustness (see \fig{fig:temporal_strength_motion_robustness}a--d).

For very low-light scenes, we have found that $c$ has to be increased considerably in order to increase the contribution of each frame to the average and to reduce noise. However, doing this introduces artifacts due to tiles erroneously being merged together (due to scene motion or alignment errors, as shown in \fig{fig:temporal_strength_motion_robustness}). In order to reduce noise in very low-light scenes without suffering from artifacts, we introduce a spatially varying scaling factor to $c$, the ``temporal strength factor'', $f_{tz}$, where $t$ and $z$ represent the tile and frame respectively. The modified temporal strength is $c \!\cdot\! f_{tz}$. We compute the temporal strength factor as a function of a mismatch map, $m_{tz}$, which is calculated per frame in the spatial (pixel) domain. The mismatch maps are calculated using a ``shrinkage'' operator \cite{dabov2007image}, with the same form as \eq{eq:HDR+merge}:
\begin{align}
    m_{tz} &= \frac{d_{tz}^2}{d_{tz}^2 + s{\sigma_{tz}}^2}\ .
    \label{eq:mismatch_map}
\end{align}
The mismatch, $m_{tz}$, is calculated between the corresponding tile in frame $z$ to tile $t$ in the reference frame, and has a value between zero (no mismatch) and one (high mismatch). $d_{tz}$ is the L1 difference between tile $t$ in the reference frame and the tile that was aligned to it from frame $z$. For efficiency, we reuse the calculations in the previous alignment stage to obtain $d_{tz}$. The constant $s$ is a scaling factor to allow tuning the mismatch maps, and we found that $s=0.5$ results in a mismatch map that suppresses noise while detecting actual motion in the scene. 

The temporal strength factor, $f_{tz}$, is a piecewise-linear function of $m_{tz}$, and the modified $A_{tz}(\omega)$ is now: 
\begin{align}
    A_{tz}(\omega) &= \frac{\abs{D_{tz}(\omega)}^2}{\abs{D_{tz}(\omega)}^2 + cf_{tz}(m_{tz}){\sigma_{tz}}^2}\ . \label{eq:modified_merge}
\end{align}
The function $f_{tz}(m_{tz})$, shown in \fig{fig:temporal_strength_motion_robustness}f, is manually tuned for various scene noise levels (and linearly interpolated between them) to increase the temporal denoising in high-matching regions while decreasing merging and preserving the details in regions that have high-mismatch. The maximal value of the temporal strength factor is higher when more merging is desired, notably, darker scenes with higher noise, as determined by a calculated tuning factor. The minimal value of the temporal strength factor is similar for all scenes and represents increased robustness to mismatch artifacts. In these cases the temporal strength factor is simply $1$ and the algorithm naturally degrades back to \eq{eq:HDR+merge}. 

This modified algorithm provides additional robustness to \cite{hasinoff2016burst} by better determining when the differences between tiles originate from changes in random noise and when they are due to actual changes in the scene. Although the the mismatch map is also based on tile differences, the difference of \eq{eq:HDR+merge}, $D_{tz}(\omega)$,  only looks at differences within a certain frequency band in a $16 \times 16$ pixel tile, while the difference $d_{tz}$ of \eq{eq:mismatch_map} originates from the alignment error which often has a larger support (up to $64 \times 64$) and is calculated in the spatial domain, across all frequency bands. These complementary approaches to calculating the differences between tiles help create a motion robust merge that is also effective at removing noise in very low light.

As an additional improvement to \cite{hasinoff2016burst}, instead of calculating the weight $1-A_{tz}(\omega)$ per color channel and merging each channel separately, we now compute the minimal weight across all color channels for each tile, and use a similar weight for merging all of the channels in each frequency band. We found that in low light the weights for each channel can be very different from one another, which causes color artifacts. Using a similar weight is important to reduce these artifacts.

\subsection{Spatially varying denoising}
\begin{figure}
    \centering
    \includegraphics[width=\linewidth]{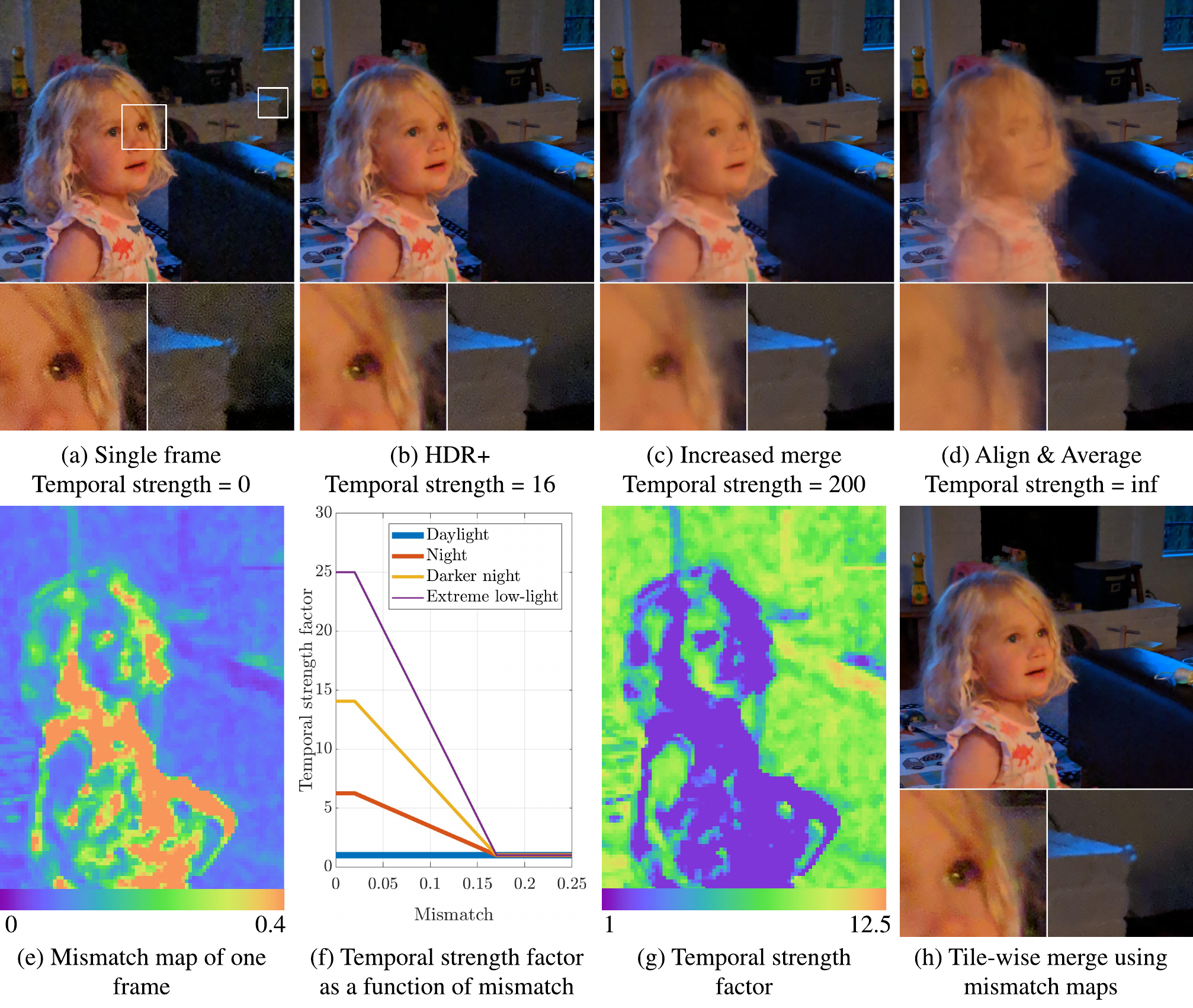}
    \caption{The temporal strength ($c$ in \eq{eq:HDR+merge}) determines the tradeoff between temporal denoising and motion robustness. It does this by adjusting the contributions of tiles in alternate frames to the merged result based on their similarity to the corresponding tiles in the reference frame. To demonstrate the influence of the temporal strength, we manually set it  to various values in sub-figures (a-d). In (a) the temporal strength is set to zero and effectively only the reference frame is used to create the image. In (b), the burst of 13 frames is merged using a low temporal strength. The resulting image has significant noise but no motion blur. In (c, d) the temporal strength is high and consequently the resulting image has less noise but more motion blur. By creating mismatch maps for each frame, we can compute a spatially varying temporal strength. (e) shows a mismatch map of one of the frames of the burst, as computed per tile according to \eq{eq:mismatch_map}. The static background has low mismatch and the regions which have motion have a high mismatch. (f) shows the conversion from the mismatch maps to the temporal strength factor at various scene noise levels. (g) shows the spatially varying temporal strength factor. In the resulting image, regions with movement, such as the face, are sharp as in (b), and the static regions, such as the corner of the fireplace, have reduced noise as in (c). Refer to Section~\ref{sec:merge_vis} of the appendix to see additional frames from this burst and the corresponding mismatch maps.}
    \label{fig:temporal_strength_motion_robustness}
\end{figure}

\begin{figure}
    \centering
    \includegraphics[width=\linewidth]{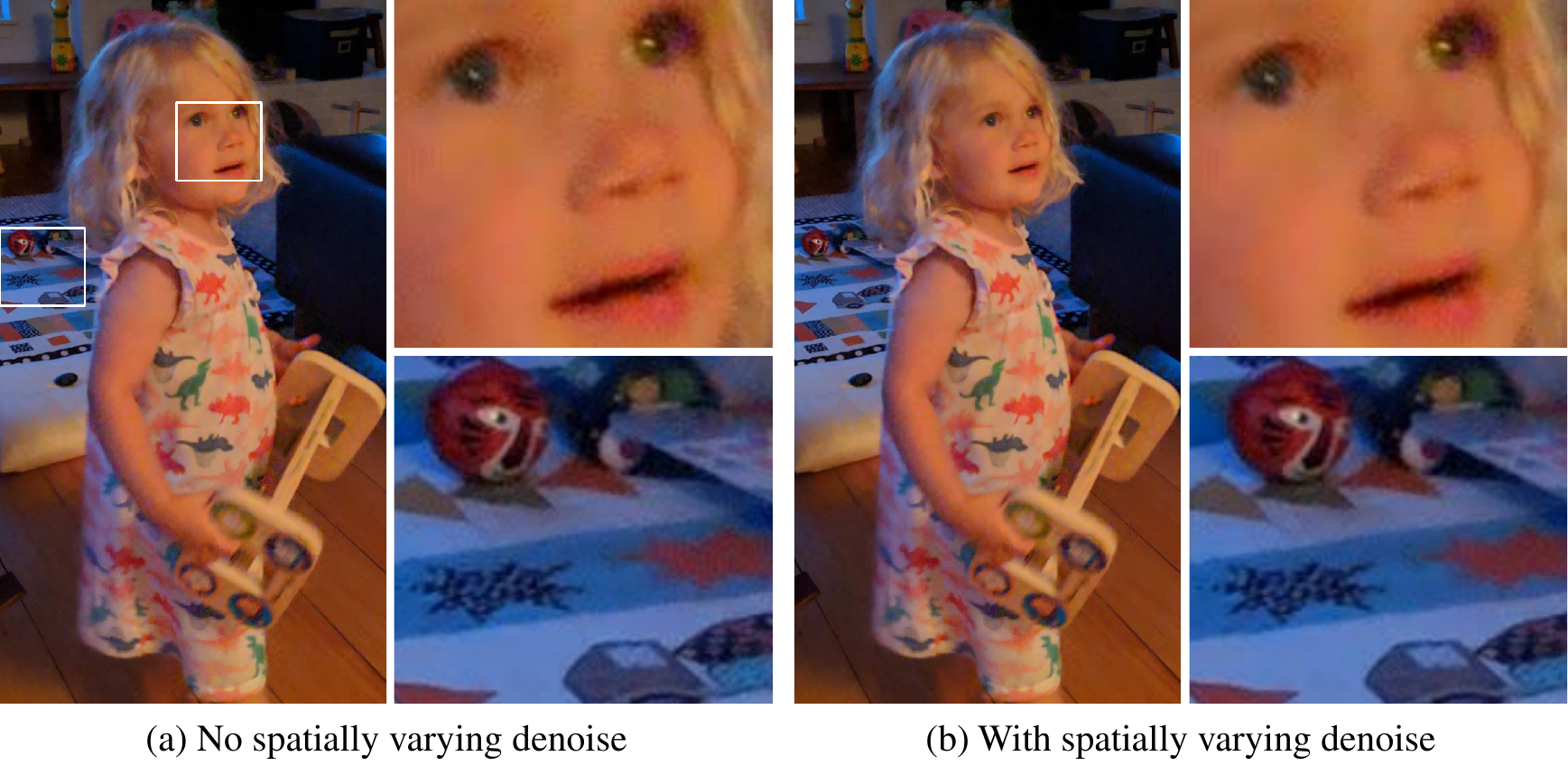}
    \caption{Spatially varying denoising adaptively reduces more noise in regions with less temporal denoising, such as the face and hand. Static regions of the image are not affected. The amount of denoising increases in tiles with reduced frame merging.}
    \label{fig:merge_w_wo_spatially_varying_denoise}
\end{figure}

As in \cite{hasinoff2016burst}, we next perform spatial denoising in the 2D DFT domain, by applying a shrinkage operator of the same form as \eq{eq:HDR+merge} to the spatial frequency coefficients. Originally, the strength of denoising was calculated assuming that all $N$ frames were averaged perfectly and the estimate of the noise variance of the image after merging was updated to be $\frac{\sigma^2}{N}$. In our modified implementation, we calculate the noise variance of the merged image based on the actual merging that occurred in each tile, which increases the noise variance in regions of high mismatch. As a result, these regions get increased spatial denoising, via the shrinkage function, which compensates for their reduced temporal denoising. See \fig{fig:merge_w_wo_spatially_varying_denoise} for the results in \fig{fig:temporal_strength_motion_robustness} without and with spatially varying denoising.

\begin{figure}
    \centering
    \includegraphics[width=\linewidth]{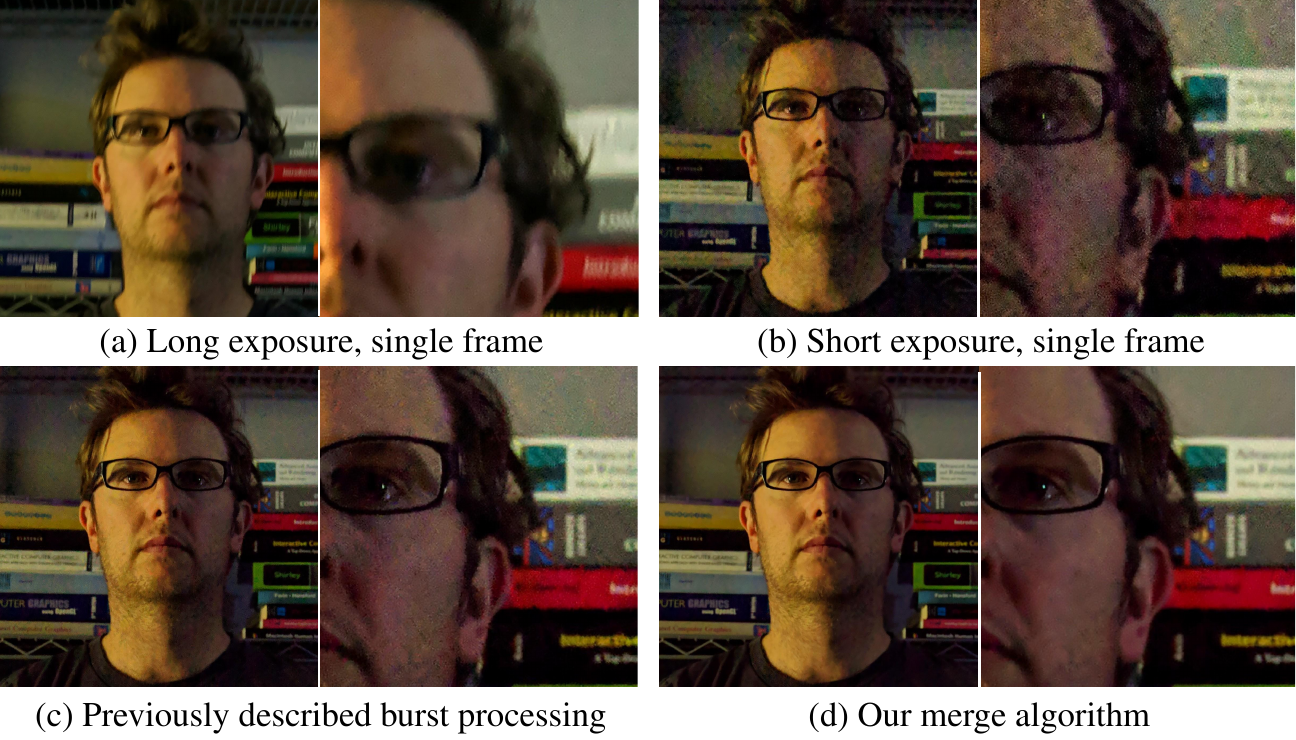}
    \caption{Motion-robust align and merge can be applied to dSLR photos to improve low light handheld photography. a) At 0.3 lux, at an exposure time of $330$\,ms at ISO 12800, the photograph isn't overly noisy on a professional dSLR (Canon EOS-1D X) but suffers from motion blur. In these situations, photographers must balance a tradeoff between unwanted motion blur and unwanted noise. b) At $50$\,ms exposure and ISO 65535 there is no motion blur, but the image suffers from significant noise. c) Merging 13 frames with $50$\,ms exposure at ISO 65535 as described in \cite{hasinoff2016burst} produces an image with reduced noise and reduced motion blur. d) When applying our modified merge algorithm to the burst we are able to reduce noise even further and provide more detail. The crops on the right were brightened by $30$\% to emphasize the differences in noise.}
    \label{fig:motion-robustness-DSLR}
\end{figure}

\fig{fig:temporal_strength_motion_robustness}h shows the result of our algorithm on a dark indoor scene with a lot of motion. The image was created by merging a burst of 13 frames captured on a mobile device. Our procedure for aligning and robustly merging multiple frames generalizes to the raw bursts captured by a handheld digital single-lens reflex camera (dSLR). As shown in \fig{fig:motion-robustness-DSLR}, our algorithm can be used to merge multiple short exposures and reduce motion blur.

%% file: 05_color.tex
\section{Low-Light Auto White Balance}

\label{sec:awb}

Camera pipelines require an Automatic White Balance (AWB) step, wherein the color of the dominant illumination of the scene is estimated and used to  produce an aesthetically pleasing color composition that (often) appears to be lit by a neutral illuminant. White balance is closely related to color constancy, the problem in both human vision \cite{foster2011color} and computer vision \cite{Gijsenij2011} of discounting the illuminant color of a scene and inferring the underlying material colors. Because modern color constancy algorithms tend to use machine learning, these algorithms can be adapted to serve as AWB algorithms by training them on illuminants that, when removed, result in pleasing color-corrected images, rather than training them on the illuminants used in color constancy that, when removed, produce veridically neutralized images.
``Fast Fourier Color Constancy'' (FFCC)~\cite{BarronTsai2017} is the current top-performing color constancy algorithm, both in terms of accuracy (on standard color constancy datasets \cite{gehler2008bayesian,Cheng14}) and speed (capable of running at ${\sim}700$ frames per second on a mobile device). FFCC works by learning and then applying a filter to a histogram of log-chrominance pixel and edge intensities, as was done by its predecessor technique \cite{BarronICCV2015}.

White balance is a particularly challenging problem in low-light environments, as the types of illuminations are more diverse, and the observed image is more noisy. In this section we describe how FFCC can be modified to serve as an effective AWB algorithm for low-light photography, through the use of a new dataset and a new error metric for training and evaluation. In Section~\ref{sec:WB_details} of the appendix we document some additional modifications we made to FFCC that are unrelated to low light photography, regarding image metadata and sensor calibration.

\subsection{Dataset}
FFCC is learning-based, and so its performance depends critically on the nature of the data used during training.
While there are publicly available color datasets \cite{gehler2008bayesian,Cheng14}, these target color constancy---objectively neutralizing the color of the illuminant of a scene. Such neutralized images often appear unnatural or unattractive in a photographic context, as the photographer often wants to retain some of the color of the illuminant in particular settings, such as sunsets or night clubs.
Existing datasets also lack imagery in extreme low-light scenarios, and often lack spectral calibration information. To address these issues, we collected 5000 images of various scenes with a wide range of light levels, all using mobile devices with calibrated sensors. 
Rather than using a color checker or gray card to recover the ``true'' (physically correct) illuminant of each scene, we recruited professional photographers to manually ``tag'' the most aesthetically preferable white balance for each scene, according to their judgment. For this we developed an interactive labelling tool to allow a user to manually select a white point for an image, while seeing the true final rendering produced by the complete raw processing pipeline of our camera. Matching the on-device rendering during tagging is critical, as aesthetic color preferences can be influenced by other steps during image processing, such as local tone-mapping, exposure, and saturation.
Our tagging tool and a subset of our training data is shown in Section~\ref{sec:WB_details} of the appendix. 

\newcommand{\error}{\Delta}
\newcommand{\angularerr}{\error_\ell}
\newcommand{\reproerr}{\error_r}
\newcommand{\weightedreproerr}{\error_{a}}
\newcommand{\ltrue}{{\boldsymbol \ell}_t}
\newcommand{\lpred}{{\boldsymbol \ell}_p}
\newcommand{\lratio}{{\boldsymbol r}}
\newcommand{\hmat}{\mathrm{H}}
\newcommand{\avgtruecol}{{\boldsymbol \mu}_t}

\subsection{Error metrics}
\begin{figure}
\begin{subfigure}[b]{.5\linewidth}
\centering
\includegraphics[width=0.98\linewidth]{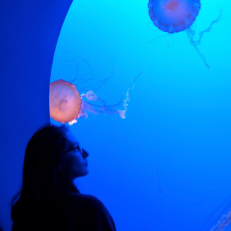}
\caption{Ground-truth image}\label{subfig:repro1_true}
\end{subfigure}%
\begin{subfigure}[b]{.5\linewidth}
\centering
\begin{overpic}[width=0.98\linewidth]{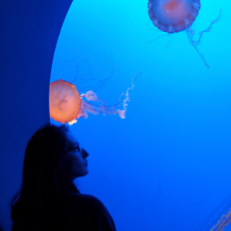}
 \put(68,23) {\colorbox{white}{\footnotesize $\angularerr = 18.1^\circ$}}
 \put(68,13) {\colorbox{white}{\footnotesize $\reproerr = 20.4^\circ$}}
 \put(68,03) {\colorbox{white}{\footnotesize $\weightedreproerr = 3.25^\circ$}}
\end{overpic}
\caption{Our model's output image }\label{subfig:repro1_ours}
\end{subfigure} \\ \vspace{0.1in}
\newcommand{\colorlosswidth}{0.93\linewidth}
\captionsetup[subfigure]{labelformat=empty}
\begin{subfigure}[b]{00.3333\linewidth}
\centering
\includegraphics[width=\colorlosswidth]{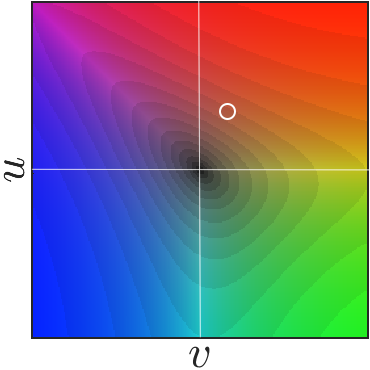}
\caption{\footnotesize \quad$\angularerr$}\label{subfig:loss_angular}
\end{subfigure}%
\begin{subfigure}[b]{0.333\linewidth}
\centering
\includegraphics[width=\colorlosswidth]{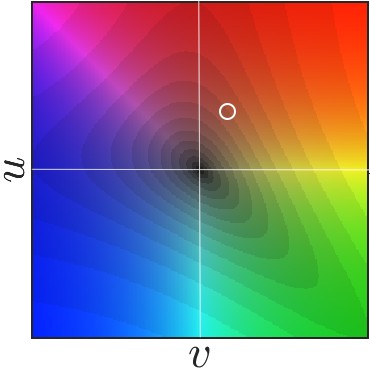}
\caption{\footnotesize \quad$\reproerr = \weightedreproerr(\avgtruecol = \vec{1})$}\label{subfig:loss_repro}
\end{subfigure}%
\begin{subfigure}[b]{0.333\linewidth}
\centering
\includegraphics[width=\colorlosswidth]{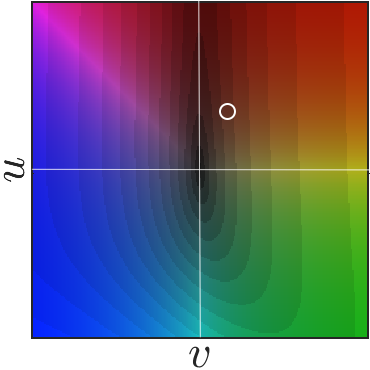}
\caption{\footnotesize \,\,\, $ \weightedreproerr( \avgtruecol = [.10; .53; .84])$}\label{subfig:loss_aqua}
\end{subfigure}
\caption{
Low-light environments often have heavily tinted illuminants, resulting in images where one or more color channels have low intensities across the image (\subref{subfig:repro1_true}). As a result, during white balance the appearance of the image is sensitive to accurate estimation of the gains for high-intensity channels (here, green and blue) but is less sensitive to the low-intensity channels (here, red), as evidenced by the output of our model (\subref{subfig:repro1_ours}), which is grossly inaccurate in the red channel but still results in an accurate-looking rendering.
Traditional error metrics ($\angularerr$ and $\reproerr$) exaggerate errors in our low-intensity color channel (\subref{subfig:repro1_ours}, inset), causing difficulties during training and evaluation. Our ``Anisotropic Reproduction Error'' (ARE), denoted $\weightedreproerr$, accounts for the anisotropy of the RGB values in the true image, and downweights errors in missing channels accordingly, resulting in a learned model that produces output images (\subref{subfig:repro1_ours}) that closely match the ground-truth (\subref{subfig:repro1_true}) in these challenging scenes.
Below we visualize the error surfaces of traditional metrics $\angularerr$ and $\reproerr$ alongside $\weightedreproerr$ using the log-chroma UV histograms of \cite{BarronTsai2017}, where luma indicates error (the true illuminant is at the origin and our prediction is a white circle). By conditioning on the mean RGB value of the true image $\avgtruecol$, the ARE's curvature is reduced with respect to red (along the $u$-axis) but still highly curved with respect to blue and green (along the $v$-axis and diagonally), which accurately reflects performance in this heavily-tinted scene.
}
\label{fig:repro}
\end{figure}

Traditional error metrics used for color constancy behave erratically in some low-light environments, which necessitates that we design our own error metric for this task.
Bright environments are often lit primarily by the sun, so the true illuminant of images taken in bright environments is usually close to white.
In contrast, low light environments are often lit by heavily tinted illuminants---campfires, night clubs, sodium-vapor lamps, etc. This appears to be due to a combination of sociological and technological trends (the popularity of colorful neon lights in dark environments) and of the nature of black body radiation (dim lights are often tinted red).
Heavily-tinted illuminants mean that the true white-balanced image may contain pixel values where a color channel's intensity is near zero for all pixels, such as the image in Fig.~\ref{subfig:repro1_true}. This is problematic when evaluating or training white balance algorithms, which are commonly evaluated in terms of how well they recover \emph{all} color channels of the illuminant. If a color channel is ``missing'' (i.e.\ having little or no intensity) then this task becomes ill-posed. For example, if all pixels in an image have a value near $0$ in their red channel, then the image will look almost identical under all possible scalings of its red channel. This is problematic for a number of reasons: 1) When collecting ground-truth illuminant data, it is unclear how missing channels of the illuminant should be set. 2) When evaluating an algorithm on heavily-tinted images, its accuracy may seem low due to irrelevant ``mistakes'' when estimating missing color channels. 3) Learning-based algorithms will attempt to minimize a loss function with respect to meaningless illuminant colors in these missing channels, which may impair learning---a model may waste its capacity trying to accurately predict illuminant colors that are impossible to determine and are irrelevant to the image's final rendering.

To address this, we present a new error metric that correctly handles heavily-tinted scenes. First, let us review the canonical metric for color constancy, which is the angular error between RGB vectors of the true illuminant $\ltrue$ and the recovered illuminant $\lpred$~\cite{Gijsenij09}:
\begin{equation}
  \angularerr\left(\lpred, \ltrue\right) = \operatorname{cos}^\mathrm{-1} \left( \frac{ \lpred^\mathrm{T} \ltrue}{ \norm{\ltrue} \norm{\lpred}} \right)
  \label{eq:angular_error}
\end{equation}
The angular error measures errors in the recovered illuminant, even though our primary concern is the appearance of the white-balanced image attained when dividing the input image by the recovered illuminant. The reproduction error of \cite{Finlayson2014} addresses this:
\begin{equation}
    \reproerr \left( \lpred, \ltrue \right) = \operatorname{cos}^\mathrm{-1} \left( \frac{ \norm{\lratio}_1 }{ \sqrt{3} \norm{\lratio}_2 } \right) \quad\quad
    \lratio = \frac{\ltrue}{\lpred}
\end{equation}
where division is element-wise. The reproduction error considers the error in the appearance of a white patch under a recovered illumination compared to the true illuminant. 
But as we have noticed, in heavily-tinted scenes where one or two color channels are missing, the notion of a ``white patch'' becomes ill-defined.
We therefore propose an improved metric that considers the appearance of the \emph{average} patch of the image when viewed under the recovered illumination, which we dub the ``Anisotropic Reproduction Error'' (ARE):
\begin{equation}
    \weightedreproerr\left(\lpred, \ltrue, \avgtruecol\right) = \operatorname{cos}^\mathrm{-1} \left( \frac{\sqrt{\lratio}^\mathrm{T} H \sqrt{\lratio}}{\sqrt{\operatorname{tr}(H)} \sqrt{  \lratio^\mathrm{T} H \lratio}} \right)
    \quad\,\,
    \hmat = \operatorname{diag}\left( \avgtruecol \right)^2
\end{equation}
$\avgtruecol$ is an RGB vector containing the mean color of the \emph{true} image:
\begin{equation}
  \avgtruecol = \underset{i}{\operatorname{mean}}\left( \begin{bmatrix} r_t^{(i)}; \,\, g_t^{(i)}; \,\, b_t^{(i)} \end{bmatrix} \right)
\end{equation}
The ARE implicitly projects the inverse of the true and predicted illuminants (i.e.\ the appearance of a white patch under the true and predicted illuminants) into a 3D space scaled by the  average color of the true white-balanced image $\avgtruecol$, and then computes the reproduction error in that projected space.
Therefore, the smaller the value of some channel of $\avgtruecol$, the less sensitive $\weightedreproerr$ is to errors in that channel of the estimated illuminant $\lpred$, with a value of $0$ in $\avgtruecol$ resulting in complete insensitivity to errors in that channel.
If two color channels are absent in an image, the ARE will be $0$ regardless of the predicted illuminant, accurately reflecting the fact that the color composition of the scene cannot be changed by applying per-channel color gains.
the ARE degrades naturally to the standard reproduction error when the average scene color is gray:
\begin{equation}
    \forall \alpha > 0, \,\, \weightedreproerr\left(\lpred, \ltrue, \alpha \bf{1} \right) = \reproerr\left(\lpred, \ltrue \right)
\end{equation}
In most of the scenes in our dataset, the average true color is close to gray and therefore the error surface of the ARE closely resembles the reproduction error, but in challenging low-light scenes the difference between the two metrics can be significant, as is shown in \fig{fig:repro}

\begin{table}

\caption{
White balance results using 3-fold cross-validation on our dataset, in which we modify FFCC by replacing its data term with different error metrics, including the ARE, $\weightedreproerr$. We report mean error, and the mean of the largest $25\%$ of errors.
Minimizing the ARE improves performance as measured by \emph{all} error metrics, as it allows training to be invariant to the inherent ambiguity of white points in heavily-tinted scenes.}
\resizebox{\linewidth}{!}{
\begin{tabular}{l|cc|cc|cc}
& \multicolumn{2}{c|}{$\angularerr$ (Ang. Err.)} & \multicolumn{2}{c|}{$\reproerr$ (Repro. Err.)} & \multicolumn{2}{c}{$\weightedreproerr$ (ARE)} \\
\multirow{2}{*}{Algorithm} & \multirow{2}{*}{Mean} & Worst & \multirow{2}{*}{Mean} & Worst & \multirow{2}{*}{Mean} & Worst \\
 & & 25\%  & & 25\% & & 25\% \\
\hline \hline
FFCC \shortcite{BarronTsai2017} &                    1.377 &                    3.366 &                    1.836 &                    4.552 &                    1.781 &                    4.370 \\
\hline
FFCC + $\reproerr$              & \cellcolor{Yellow} 1.332 & \cellcolor{Yellow} 3.154 & \cellcolor{Yellow} 1.771 & \cellcolor{Yellow} 4.255 & \cellcolor{Yellow} 1.721 & \cellcolor{Yellow} 4.088 \\
FFCC + $\angularerr$            & \cellcolor{Orange} 1.320 & \cellcolor{Orange} 3.115 & \cellcolor{Orange} 1.768 & \cellcolor{Orange} 4.229 & \cellcolor{Orange} 1.714 & \cellcolor{Orange} 4.049 \\
FFCC + $\weightedreproerr$      & \cellcolor{Red}    1.312 & \cellcolor{Red}    3.088 & \cellcolor{Red}    1.752 & \cellcolor{Red}    4.194 & \cellcolor{Red}    1.696 & \cellcolor{Red}    4.007 \\
\end{tabular}
}

\label{table:colorresults}
\end{table}
In addition to its value as a metric for measuring performance, the ARE can be used as an effective loss during training.
To demonstrate this we trained our FFCC model on our dataset using the same procedure as was used in \cite{BarronTsai2017}: three-fold cross validation, where hyperparameters are tuned to minimize the ``average'' error used by that work.
We trained four models: a baseline in which we minimize the von Mises negative log-likelihood term used by \cite{BarronTsai2017}, and three others in which we replaced the negative log-likelihood with three error metrics: $\reproerr$, $\angularerr$, and our $\weightedreproerr$ (all of which are differentiable and therefore straightforward to minimize with gradient descent).
Results can be seen in Table~\ref{table:colorresults}.
As one would expect, minimizing the ARE during training produces a learned model that has lower AREs on the validation set.
But perhaps surprisingly, we see that training with ARE produces a learned model that performs better on the validation data \emph{regardless of the metric used for evaluation}.
That is, even if one only cared about minimizing angular error or reproduction error, ARE is still a preferable metric to minimize during training.
This is likely because the same ambiguity that ARE captures is also a factor during dataset annotation: If an image is lacking some color, the annotation of the ground-truth illuminant for that color will likely be incorrect.
By minimizing ARE, we model this inherent uncertainty in our annotation and are therefore robust to it, and our learned model's performance is not harmed by potentially-misleading ground truth annotations.

%% file: 06_tone_mapping.tex
\section{Tone Mapping}

\label{sec:tone_mapping}

The process of compressing the dynamic range of an image to that of the output is known as ``tone mapping''. In related literature, this is typically achieved through the application of tone mapping operators (TMOs), which differ in their intent: some TMOs try to stay faithful to the human visual system, while others attempt to produce images that are subjectively preferred by artistic experts. \cite{ledda2005evaluation,eilertsen2016evaluation} provide evaluations of TMOs for various intents.

\begin{figure}
    \begin{center}
    \includegraphics[width=\linewidth]{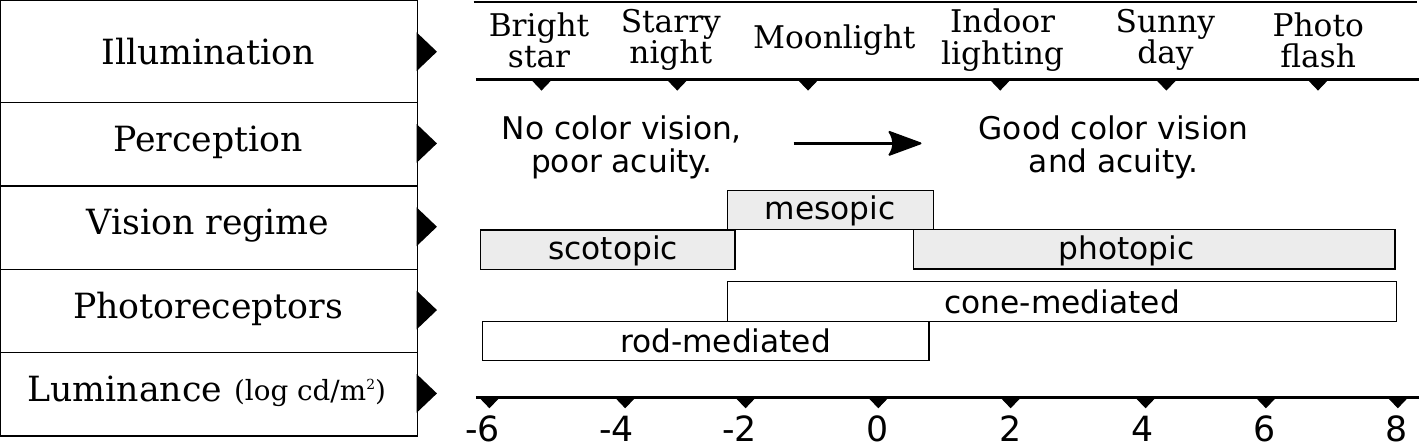}
    \end{center}
    \caption{The variation in color perception and acuity of human vision across typical ambient light levels (figure adapted from \cite{stockman2006into}). Any TMO that attempts to simulate the human visual system in low-light situations (scotopic and mesopic regimes) is subject to complex physiological phenomena, including Purkinje shift, desaturation, and loss of spatial acuity~\cite{shin2004change}.}
    \label{fig:human_vision_regimes}
\end{figure}

\fig{fig:human_vision_regimes} depicts the human visual acuity in different light levels. The intent of our TMO is to render images that are vibrant and colorful even in low-light scenarios, enabling photopic-like vision at mesopic and scotopic light levels. While TMOs are able to simulate the human visual system in the color-sensitive cone mediated regime, simulating the rod-mediated regime is more challenging due to the presence of complex physiological phenomena~\cite{ferwerda1996model,kirk2011perceptually,Jacobs:2015:SVE:2774971.2714573}. The ability to render vibrant and colorful dark scenes has been developed for digital astronomy~\cite{villard2002creating}. Our system aims to bring this ability to handheld mobile cameras.

Development of the final look of a bright nighttime photograph needs to be done carefully. A na\"ive brightening of the image can lead, for example, to undesired saturated regions and low contrast in the result. Artists have known for centuries how to evoke a nighttime aesthetic through the use of darker pigments, increased contrast, and suppressed shadows (\cite{helmholtz1995relation}, originally written in 1871). In our final rendition, we employ the following set of carefully tuned heuristics atop the tone mapping of \cite{hasinoff2016burst} to maintain a nighttime aesthetic:
\begin{itemize}
    \item Allow higher overall gains to render brighter photos at the expense of revealing noise.
    \item Limit boosting of the shadows, keeping a significant amount of the darkest regions of the image near black to maintain global contrast and preserve a feeling of nighttime.
    \item Allow compression of higher dynamic ranges to better preserve highlights in challenging conditions, such as night scenes with light sources.
    \item Boost the color saturation in inverse proportion to scene brightness to produce vivid colors in low-light scenes.
    \item Add brightness-dependent vignetting.
\end{itemize}

\cite{hasinoff2016burst} uses a variant of synthetic exposure fusion~\cite{mertens2007exposurefusion} that splits the full tonal range into a shorter exposure (``highlights'') and longer exposure (``shadows''), determines the gains that are applied to each exposure, and then fuses them together to compress the dynamic range into an 8-bit output. For scenes with illuminance $E_v$ below log-threshold $L_{\mathrm{max}}$, we gradually increase the gain $A_s$ applied to the shadows (the longer synthetic exposure) with the lux level, reaching a peak of over 1 additional stop (2.2$\times$) of shadow brightness at log-threshold $L_{\mathrm{min}}$:
\begin{equation}
  A_s = 2.2^{1 - \max \left(0, \min \left(1, \frac{\operatorname{log}(E_v) - L_{\mathrm{min}}}{L_{\mathrm{max}} - L_{\mathrm{min}}} \right) \right) }\ ,
  \label{eq:tone_mapping_shadow_gain}
\end{equation}
where the illuminance thresholds $L_{\mathrm{max}}$ and $L_{\mathrm{min}}$ correspond to a scene brightness of 200 lux and 0.1 lux, respectively. 

A fraction of up to 20\% of the additional shadow gain $A_s$ is also applied to the highlights (the short synthetic exposure). This is done in inverse proportion to the normalized dynamic range of the scene $D$, because otherwise the intensity distribution can become too narrow, causing a loss of contrast. The highlight gain $A_h$ is:
\begin{equation}
\begin{aligned}
  A_h &= 1 + 0.2 \cdot (A_s - 1) \cdot (1 - D)\ .
  \label{eq:tone_mapping_highlight_gain}
\end{aligned}
\end{equation}
We also increase the color saturation in inverse proportion to the log-lux level of the scene up to $20$\%, which helps maintain vivid renditions of facial skin tones.

Due to the high gains applied to the images, artifacts from inaccuracies in the subtraction of the black-level and lens-shading correction begin to manifest. These artifacts are most prominent in the corners of the image (where the lens-shading correction gain is greatest). To suppress these artifacts, we apply vignetting in scenes darker than 5 lux, gradually increasing in darker scenes. The appearance of vignetting also contributes to the perception of a dark scene and adds an aesthetic ``depth'' to the image.

\begin{figure}
    \centering
    \begin{subfigure}[b]{0.31\columnwidth}
        \centering
        \includegraphics[width=\textwidth]{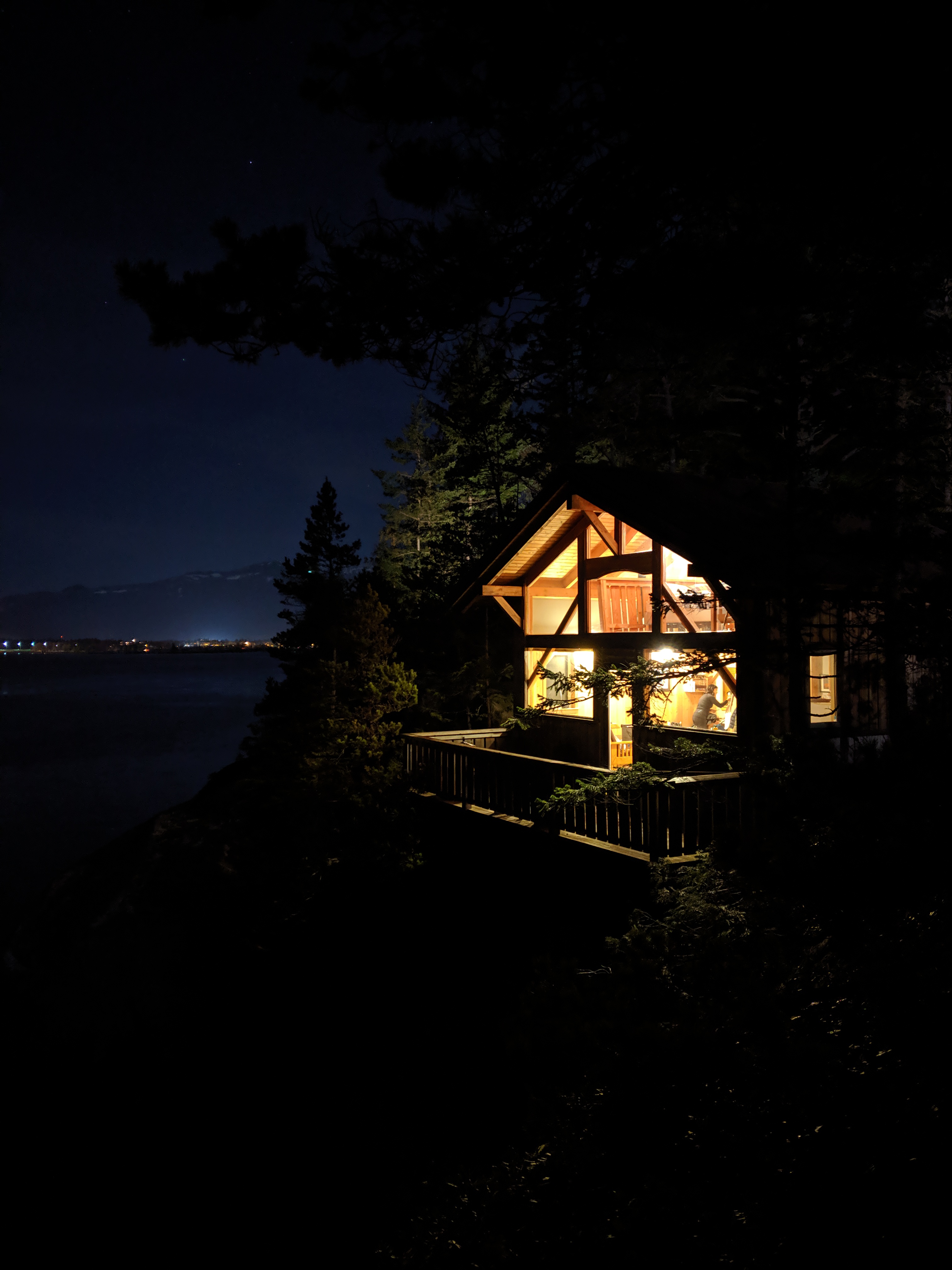}
        \caption{Baseline}
    \end{subfigure}
    \begin{subfigure}[b]{0.31\columnwidth}
        \centering
        \includegraphics[width=\textwidth]{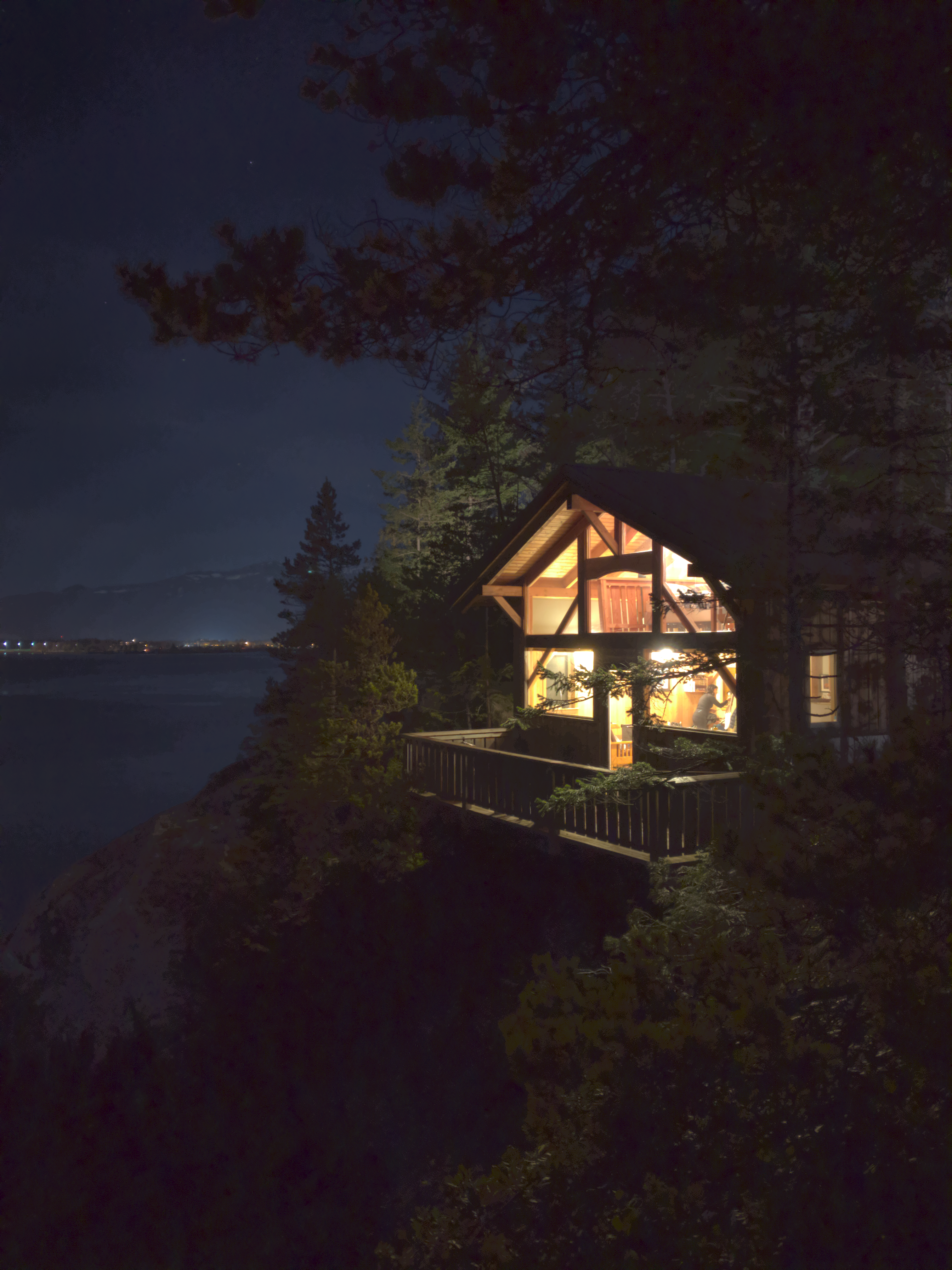}
        \caption{CLAHE}
    \end{subfigure}
    \begin{subfigure}[b]{0.31\columnwidth}
        \centering
        \includegraphics[width=\textwidth]{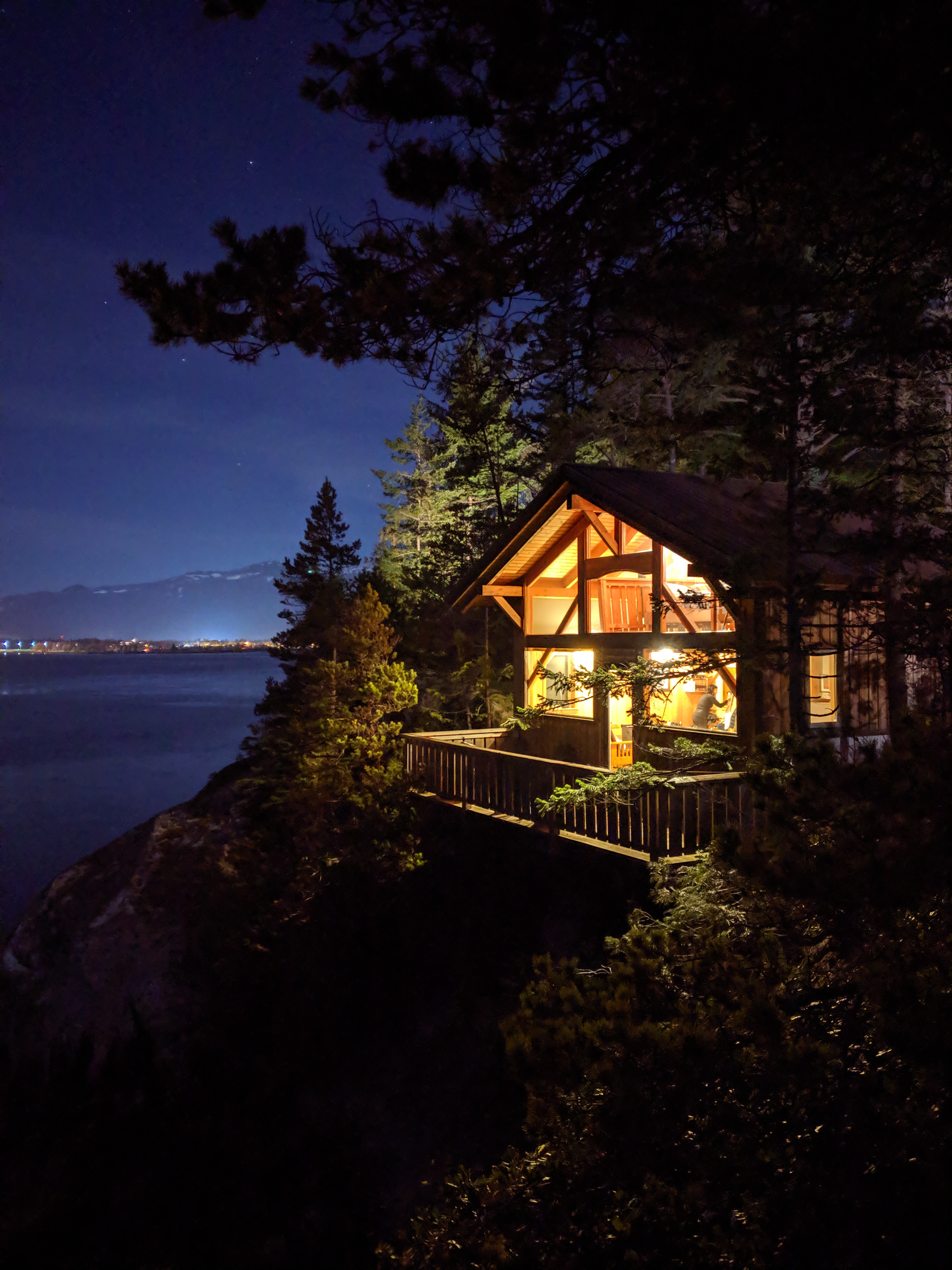}
        \caption{Our tone mapping}
    \end{subfigure}
    \caption{A nighttime scene (about 0.4 lux) that demonstrates the difficulty in achieving balanced tonal quality. (a) The tone mapping of \cite{hasinoff2016burst} results in most of the image being too dark and is not a pleasing photograph. (b) Applying a tone mapping technique that brightens the image using histogram equalization (CLAHE~\cite{zuiderveld1994contrast}) reveals more detail, but the photo lacks global contrast. (c) The photo with our tone mapping retains the details of (b), but contains more global contrast and enough dark areas to show that the captured scene is dark.}
    \label{fig:tonemap_comparison}
\end{figure}

To demonstrate the results of our method, the rendition from~\cite{hasinoff2016burst} is shown in~\fig{fig:tonemap_comparison}a and is too dark to show much detail. The amount of detail in that rendition can be increased by locally increasing its contrast, for example by applying adaptive histogram equalization. In~\fig{fig:tonemap_comparison}b we use CLAHE~\cite{zuiderveld1994contrast} on the V channel following an HSV decomposition of the image \cite{smith1978color}. This increases local contrast but the global contrast is flattened, resulting in a low-contrast appearance. The result of our technique (\fig{fig:tonemap_comparison}c) reveals shadow details without significant loss of contrast and maintains a natural nighttime appearance.

%% file: 07_results.tex
\section{Results}
\label{sec:results}

Our system was launched publicly in November 2018 as a mode called ``Night Sight'' in the camera app on Google Pixel smartphones. It is implemented on Android and relies on the Camera2 API \cite{camera2}. On Pixel, Pixel 2 and Pixel 3a we use the motion-adaptive burst merging method described in \sect{sec:motion_robustness}. On Pixel 3, which has a faster system-on-chip, burst merging is performed using \cite{sabre2019}. This alternative merging technology also launched in October 2018 on the Pixel 3 under the name ``Super Res Zoom''. We compare these two merging algorithms in Section~\ref{sec:merge_SR} of the appendix. All of the results in this paper, except for those clearly indicated in \fig{fig:sabre_comparison}, use the motion-robust Fourier-domain merge described in \sect{sec:motion_robustness}.

Our system uses refined and tuned versions of some of the algorithms described in the ``Finishing'' section of \cite{hasinoff2016burst} in order to convert the raw result of the merged burst into the final output. These include demosaicing, luma and chroma spatial denoising, dehazing, global tone adjustment, and sharpening. We are not able to provide additional details about these proprietary algorithms, and we note that they are not a contribution of the system described in this paper.

\begin{figure*}
    \centering
    \includegraphics[width=\textwidth]{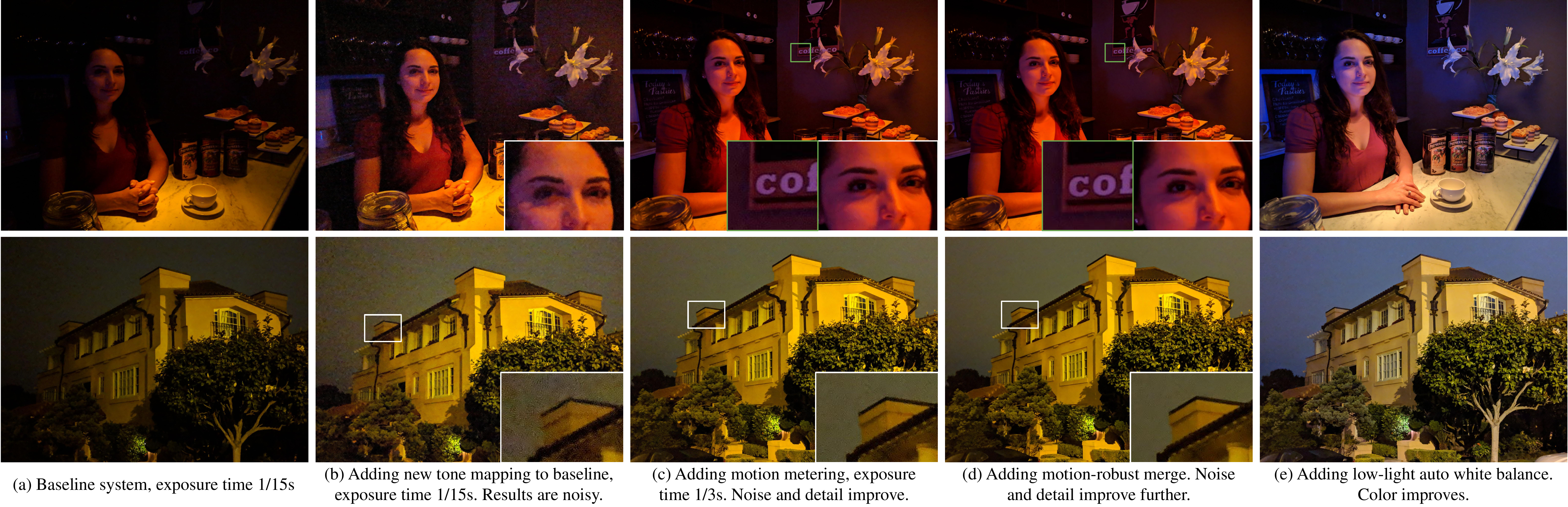}
    \caption{Comparison between our pipeline and the baseline burst processing pipeline~\cite{hasinoff2016burst} on 13 frame bursts. Due to hand shake, over the duration of the capture the top example contains 41 pixels of motion and the bottom example contains 66 pixels of motion. The columns show the effect of augmenting the baseline with various parts of our pipeline. (a) shows the results of the baseline. (b) demonstrates how our tone mapping makes the scene brighter, allowing the viewer to see more detail while preserving global contrast. However, noise becomes more apparent, and overwhelms visible detail in the darker parts of the scenes. Adding motion metering in (c) allows us to increase the exposure time in these scenes without introducing significant motion blur (3-4 pixels per frame). The resulting images are less noisy, and more detail is visible. Adding motion-robust merging in (d) allows us to increase temporal denoising in static regions, further improving detail. Finally, the low-light auto white balance is applied in (e), making the colors in the scene better match the perceived colors from the human visual system.}
    \label{fig:results_hdrp}
\end{figure*}

In \fig{fig:results_hdrp}, we compare our results to the burst photography pipeline that we build upon~\cite{hasinoff2016burst}. The sequence shows that simply applying the tone mapping described in \sect{sec:tone_mapping} to the baseline results in excessive noise. When motion metering (\sect{sec:motion_metering}) and motion-adaptive burst merging (\sect{sec:motion_robustness}) are applied, the noise and detail improve significantly. Finally, the low-light auto white balance (\sect{sec:awb}) algorithm improves color rendition, making the final image better match the human visual perception of the scene. The final results allow users to see more detail while still having low noise, pleasing colors, and good global contrast. In \fig{fig:wall_of_images}, we demonstrate our processing compared to~\cite{hasinoff2016burst} across multiple scenes and a variety of illuminants. Additionally, we have processed the dataset released by \cite{hasinoff2016burst} (\cite{hdrplusdata}) and published our results on \cite{NSWebsite}.

\begin{figure*}
    \centering
    \includegraphics[width=\textwidth]{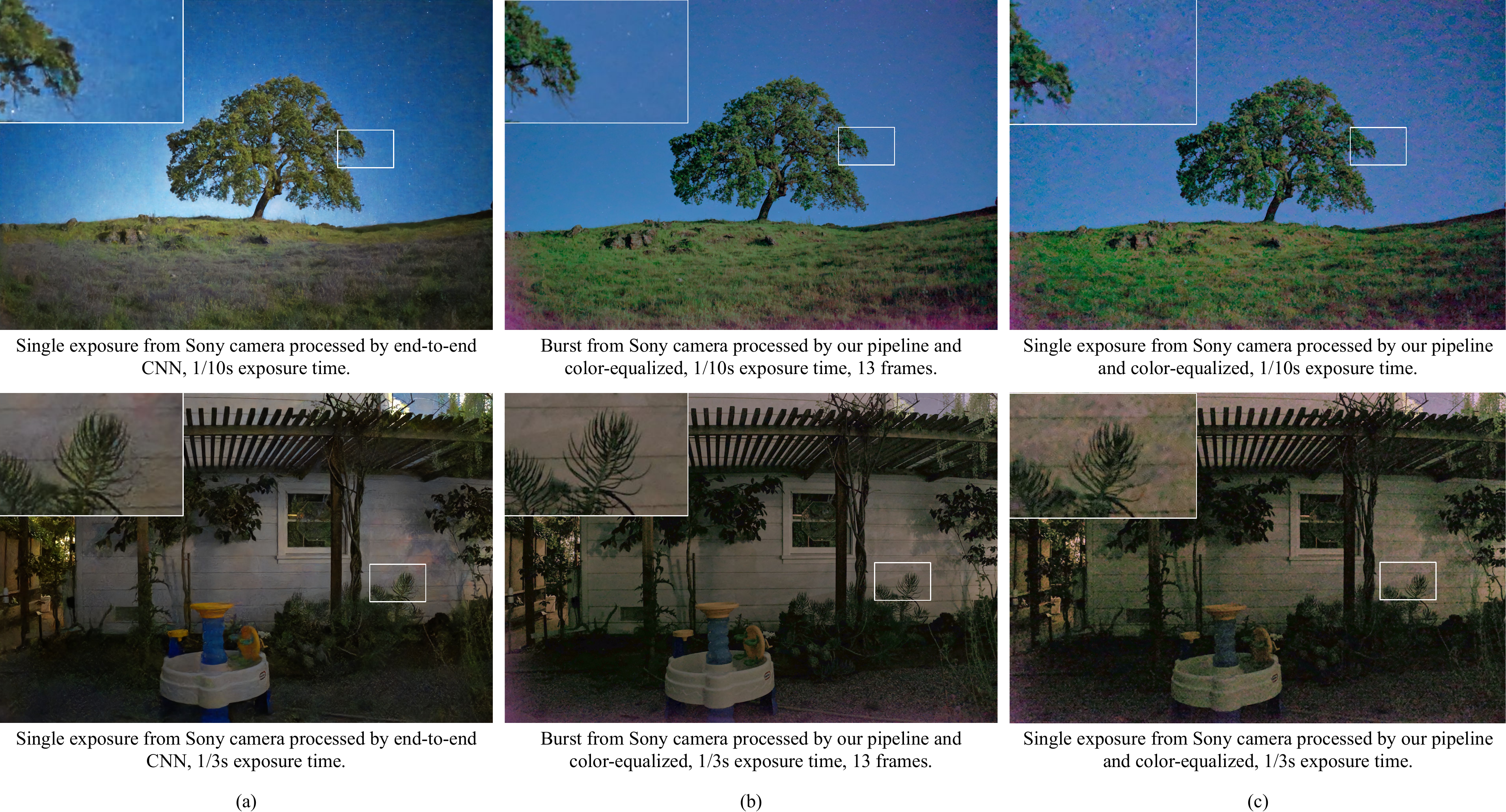}
    \caption{Comparison between the end-to-end trained CNN described in \cite{chen2018learning} and the pipeline in this paper. Raw images were captured with a Sony $\alpha$7S II, the same camera that was used to train the CNN. The camera was stabilized using a tripod. The illuminance of these scenes was 0.06-0.15~lux. Because our white balancing algorithm was not calibrated for this sensor, and since color variation can affect photographic detail perception, the results of our pipeline are color-matched to the results of \cite{chen2018learning} using Photoshop's automatic tool (see \fig{fig:results_l2sitd_color_correction} of the appendix for details). Column (a) shows the result from the CNN, and column (b) shows the result of our pipeline processing 13 raw frames. Column (c) shows the result of our pipeline processing one raw frame (no merging). The advantages of burst processing are evident, as the crops in column (b) show significantly more detail and less noise than the single-frame results. Additional examples can be found in \fig{fig:results_l2sitd_supp} of the appendix.}
    \vspace{-0.2in}
    \label{fig:results_l2sitd}
\end{figure*}

We also compare our pipeline with an end-to-end trained neural network that operates on raw images~\cite{chen2018learning}. The CNN performs all of the post-capture tasks performed by our pipeline, but with a large convolutional architecture and a single frame as input. \fig{fig:results_l2sitd} and \fig{fig:results_l2sitd_supp} show that the output of the CNN has less detail and more artifacts than the output of our pipeline. In this comparison, we used raw frames captured with a similar camera model as was used for training the network in~\cite{chen2018learning} (Sony $\alpha$7S II) because that network did not generalize well to the frames of the mobile device and produced severe color and detail artifacts (not shown here). Although our pipeline was not specifically calibrated or tuned for the Sony camera, it still produced higher-quality images compared with the results of~\cite{chen2018learning}.

Performance on mobile devices must also be considered while evaluating imaging pipelines. The CNN network architecture in~\cite{chen2018learning} is large and may not be well suited for running efficiently on a mobile device. \cite{chen2018learning} reports a time of $0.38$\,s to process the Sony images on a desktop GPU, which is equivalent to $10$s of seconds on a mobile GPU \cite{GFXBench} (assuming memory and temperature limits allow the model to be run on a mobile device). Our pipeline has been designed for efficiency, and thus is able to process the captured burst of images in under 2 seconds on a mobile phone. Pre-capture motion metering is also efficient and can estimate the flow field's magnitude in only $3$\,ms. Therefore, the maximum total capture and processing time of our system is $8$\,s, of which $6$\,s is the maximum capture time, achieved when the camera is stabilized. Memory is also an important consideration. Based on the architecture of the network described in \cite{chen2018learning}, we estimate that it requires around $3$\,GB of memory, while our process needs under $400$\,MB of memory. 

\begin{figure}
    \centering
    \includegraphics[width=\linewidth]{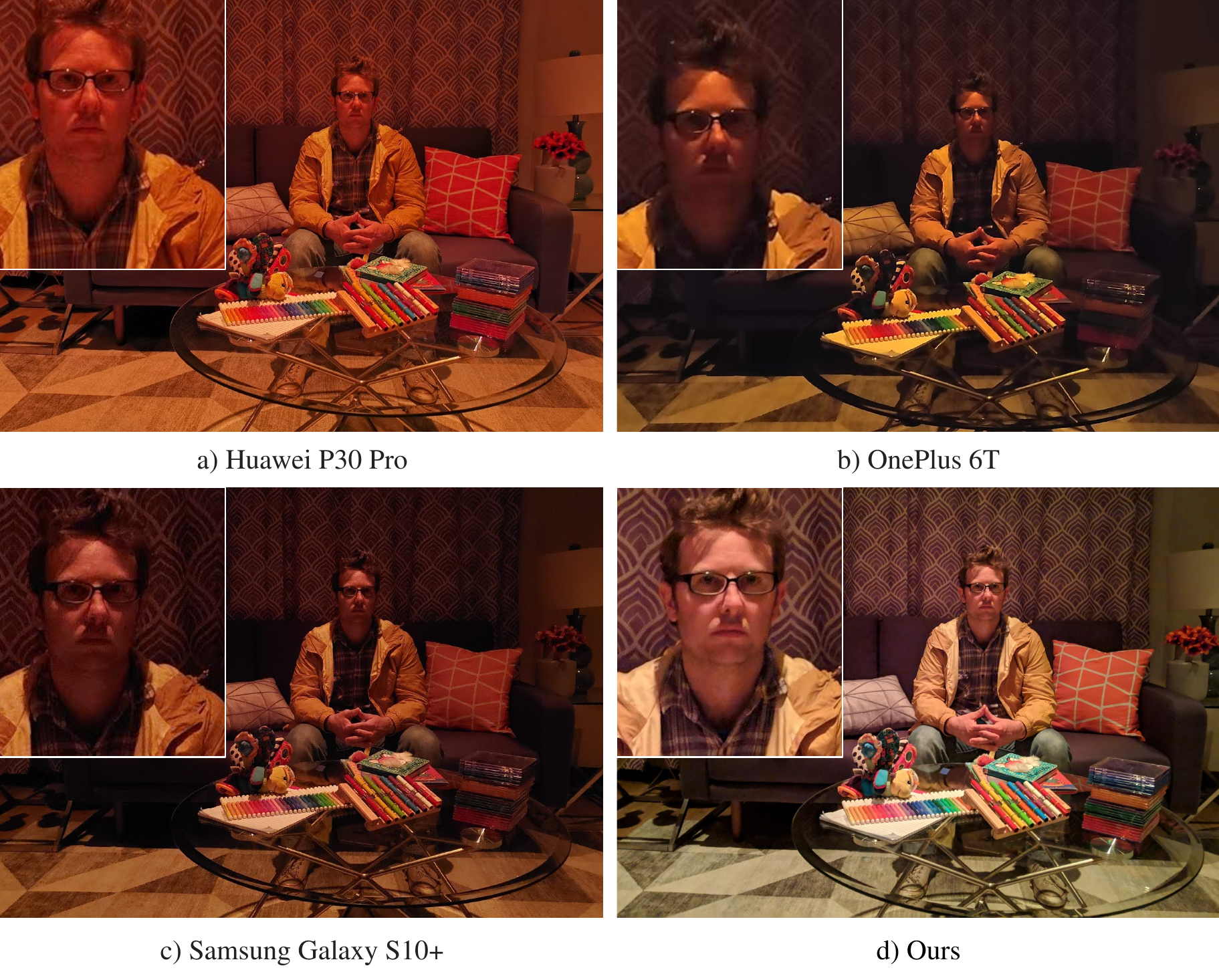}
    \caption{Comparison of the output of our system to photographs captured with the specialized low-light modes of other devices. The scene brightness is 0.4 lux and it is illuminated by a dim tungsten light. Readers are encouraged to zoom in on the images. The different results show the challenges inherent in low-light photography. High levels of detail, limited artifacts, pleasing color, and good exposure (brightness) are all important, and are difficult to achieve in unison.}
    \label{fig:different_device_comparison}
\end{figure}

\fig{fig:different_device_comparison} compares photos captured with our device and processed with our on-device pipeline, to three other devices that have specialized low-light photography modes. Our pipeline has a well-balanced rendition that has good brightness and pleasing color with a low amount of artifacts. For additional examples, please refer to Section~\ref{sec:results_appendix} of the appendix. 

\subsection{Limitations} 

\label{sec:limitations_future_work}

Although our proposed system is successful in producing high-quality photographs on a mobile device in low-light conditions, it has several practical limitations.

\textbf{Autofocus.} Consistent and accurate autofocus (AF) is a prerequisite for sharp output images. AF in low-SNR scenes is fundamentally difficult, and beyond the scope of this paper. We use the device's AF out of the box, which employs both phase- and contrast-based strategies, but isn't optimized for very low-light conditions. Empirical experiments show that our camera's AF fails almost $50\%$ of the time when the scene luminance is around 0.4~lux. We mitigate this issue by locking the lens's focus distance to the hyperfocal distance when the AF system reports a focus failure. When the lens is at the hyperfocal distance, everything from a certain distance out to infinity is in acceptable focus, and the depth of field is maximized, thereby minimizing potential misfocus. We also allow the user to adjust the focus distance manually prior to the capture.

\textbf{Real-time preview.} The viewfinder allows the user to frame the shot in real time and anticipate the final rendition of a processed image. Rendering a high-quality preview in low light is challenging, as both exposure time and processing latency are constrained by the viewfinder refresh rate (e.g., $\geq$ $15$\,fps). We circumvent this challenge by adopting the camera's proprietary hardware-accelerated preview pipeline and adaptively gaining up its output using a GPU shader. One could potentially obtain better results by employing lightweight temporal denoising techniques such as \cite{llv}.

\textbf{Shutter lag.} Similar to \cite{hasinoff2016burst}, our system doesn't start collecting raw frames until the user presses the shutter, resulting in a noticeable delay for the captured ``moment''. This is different from many zero-shutter-lag (ZSL) applications, where the camera continuously caches viewfinder frames to a ring buffer and processes them upon shutter press. We choose not to use ZSL frames because their exposure time is limited by the viewfinder refresh rate, which is so short that read noise dominates the signal.

\textbf{Black levels.} The minimum digital value returned by a sensor that is treated as black is called the black level. Inaccurate black levels are a potential limitation to low-light imaging. However, a good sensor can get within 0.25 data numbers (DN) of a 10-bit signal, and accuracy can be further improved with a gain-dependent look-up table. We use black levels straight from the camera driver.

\textbf{Sky tones.} At night, the tone of the sky usually overlaps with that of the foreground. As our pipeline adjusts the mid-tones during global tone mapping to make details in the foreground more visible, the rendition of the sky is also brightened. The perception of darkness is naturally sensitive to the tone and color of the sky and therefore in certain conditions our results fail to convey the sense of darkness adequately. To correctly account for the perception of darkness, one would need to apply a different tonal adjustment specifically to the sky.

%% file: 08_conclusion.tex
\section{Conclusions}

\label{sec:conclusions}

In this paper, we set forth a list of requirements that a computational photography system should satisfy in order to produce high-quality pictures in very low-light conditions, and described an implementation of a system that meets these requirements. We introduced motion metering, a novel approach for determining burst capture parameters that result in a good tradeoff between SNR and motion blur, and motion-robust merging, which minimizes motion artifacts when merging extremely noisy raw images. We also outlined how we incorporate learning-based auto white balancing and specialized tone mapping to best convey the color and aesthetics of a low-light photograph. Our system overcomes many constraints of a small sensor size, and runs efficiently on a modern mobile device.

%% file: 09_acknowledgement.tex
\begin{acks}
\label{sec:acknowledgements}

Night Sight and the algorithms presented in this paper would not have been possible without close collaboration with Google Research's Gcam team, Luma team, and Android camera team. From Gcam we particularly thank David E. Jacobs, Zalman Stern, Alexander Schiffhauer, Florian Kainz, and Adrian Rapazzini for their help on the project and paper, and staff photographers Michael Milne, Andrew Radin, and Navin Sarma. From the Luma team, we particularly thank Peyman Milanfar, Bartlomiej Wronski, and Ignacio Garcia-Dorado for integrating and tuning their super-resolution algorithm for Night Sight on Pixel 3 and for their helpful discussions.
\end{acks}

%% file: references.bbl

%% file: supplemental_text.tex
\appendix

\ExplSyntaxOn
\newcommand\latinabbrev[1]{
  \peek_meaning:NTF . {
    #1\@}
  { \peek_catcode:NTF a {
      #1.\@ }
    {#1.\@}}}
\ExplSyntaxOff
\def\etal{\latinabbrev{et al}}

\section{Visualization of Motion-adaptive burst merging}
\label{sec:merge_vis}

\begin{figure}

    \centering
    \vspace{0.1in}
    \includegraphics[width=\linewidth]{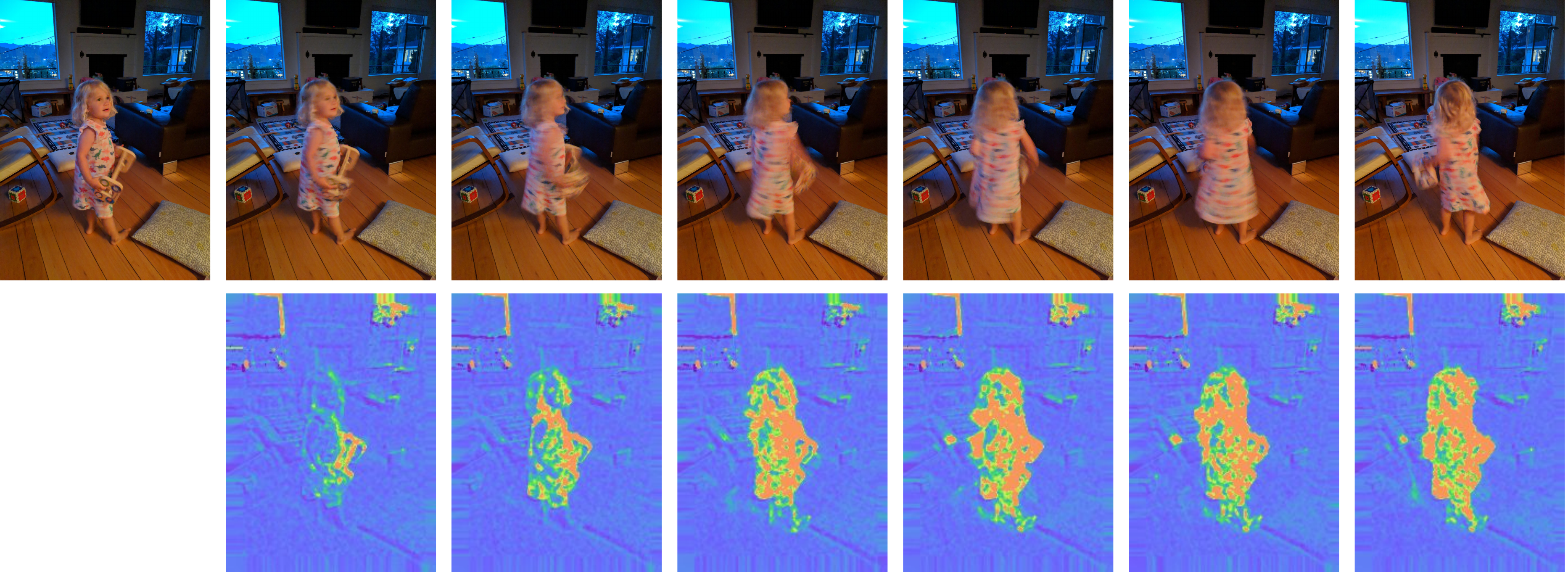}
    \caption{Seven out of the thirteen frames in the burst that was used to demonstrate motion-robust merge in Figure 9 of the main text. The bottom row shows the mismatch maps, which were calculated with the first frame as the reference frame.}
    \label{fig:motion_in_burst}
\end{figure}

\fig{fig:motion_in_burst} shows extensive motion in 7 out of the 13 frames of the burst that was used to demonstrate motion-robust merge in Figure 9 of the main text. To create the images in \fig{fig:motion_in_burst}, we processed the burst with our described algorithm, only changing the reference frame from the first to the seventh. The mismatch maps were created with the first frame as the reference frame.

\begin{figure}
    \centering
    \includegraphics[width=\linewidth]{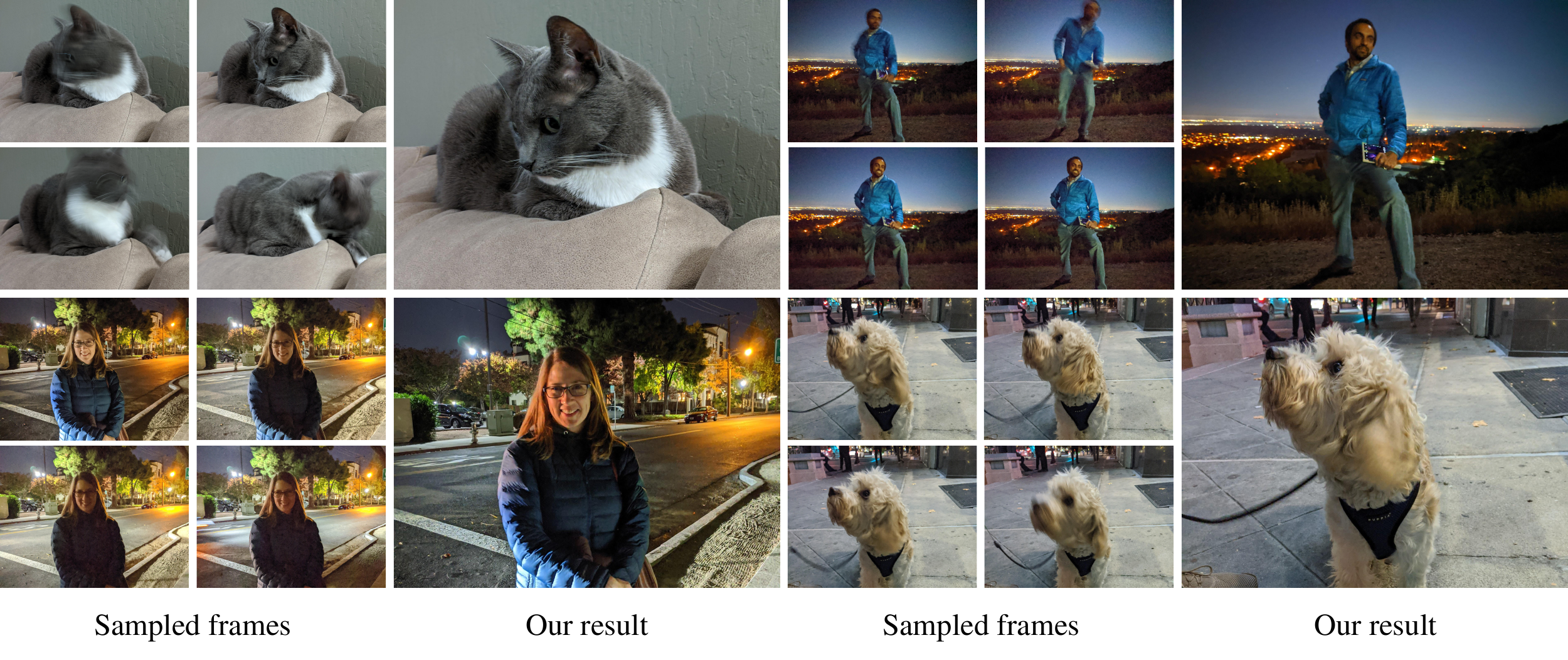}
    \caption{Scenes with fast changing objects or illuminants and the result of our system. The frames were uniformly sampled from the the 13 frames of the burst.}
    \label{fig:motion_bursts}
\end{figure}

In \fig{fig:motion_bursts}, we show a sample of scenes with fast motion or changing illumination and the results of our system.

\begin{figure}[b!]
    \centering
    \includegraphics[width=\linewidth]{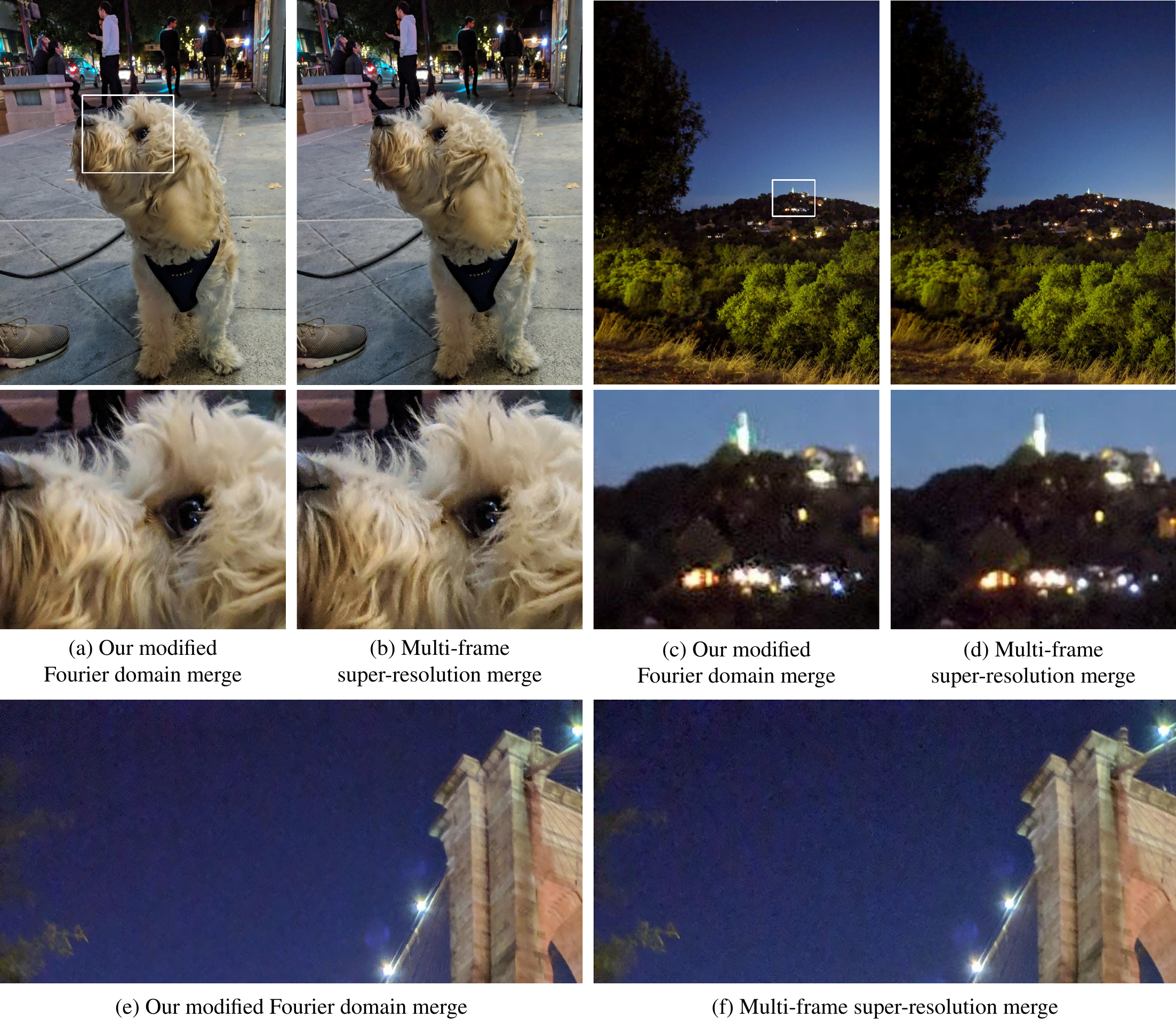}
    \caption{Merge and denoise in the frequency domain (ours) compared to the spatial domain (\cite{sabre2019}). While the spatial domain algorithm produces sharper results in the presence of motion, it is more computationally expensive than our technique. (a), (b) The spatial domain algorithm preserves detail better than the frequency domain algorithm. (c), (d) The frequency domain algorithm produces artifacts around bright lights which do not appear in the spatial-domain algorithm. (e), (f) The frequency-domain algorithm is better at reducing low-frequency noise compared to the spatial-domain algorithm.}
    \label{fig:sabre_comparison}
\end{figure}

\section{Merging using motion-robust super-resolution}
\label{sec:merge_SR}

Multi-frame super-resolution algorithms and temporal merging to reduce noise both require steps to align and merge a burst of frames, as well as being robust to scene motion. An alternative to merging in the Fourier domain is the recently developed motion-robust super resolution algorithm for handheld devices \cite{sabre2019}. In this work, the burst is merged directly onto an RGB grid in the spatial domain by accumulating frame contributions using locally adaptive kernels. Use of locally adaptive merge kernels enables adjustment of the strength of local denoising, depending on the presence of local features like edges or textures. This technique achieves local robustness to local motion, misalignment and occlusions through local statistical analysis of the merged frames, which is analogous to the mismatch maps described in Section~\ref{sec:motion_robustness}. Accurate per-pixel information about the number of locally merged frames allows it to guide a spatial denoiser to be more aggressive in regions with less temporal merging. Owing to its per-pixel, rather than per-tile, robustness and denoise calculations and merge strategy, \cite{sabre2019} is generally able to show more spatial detail in regions like faces and textures (\fig{fig:sabre_comparison} (a, b)) and provide better high frequency motion robustness. However, it requires a specialized implementation suited to the target GPU hardware and operating system, which may not be applicable on all mobile devices.

An additional difference between the two methods is that the spatial-domain merge and denoise can reproduce sharp bright light sources more accurately. By comparison, denoising in the Fourier domain can cause typical ringing artifacts (\fig{fig:sabre_comparison} (c, d)). On the other hand, implementing denoising in the Fourier domain is more effective in very low-light scenes, where noise is more dominant. This can result in better perceptual quality of the noise texture in uniform regions such as skies (\fig{fig:sabre_comparison} (e, f)).

\section{White Balance Details}
\label{sec:WB_details}

\subsection{Dataset}
\begin{figure}[b!]
    \centering
    \begin{subfigure}[b]{\linewidth}
       \includegraphics[width=\linewidth]{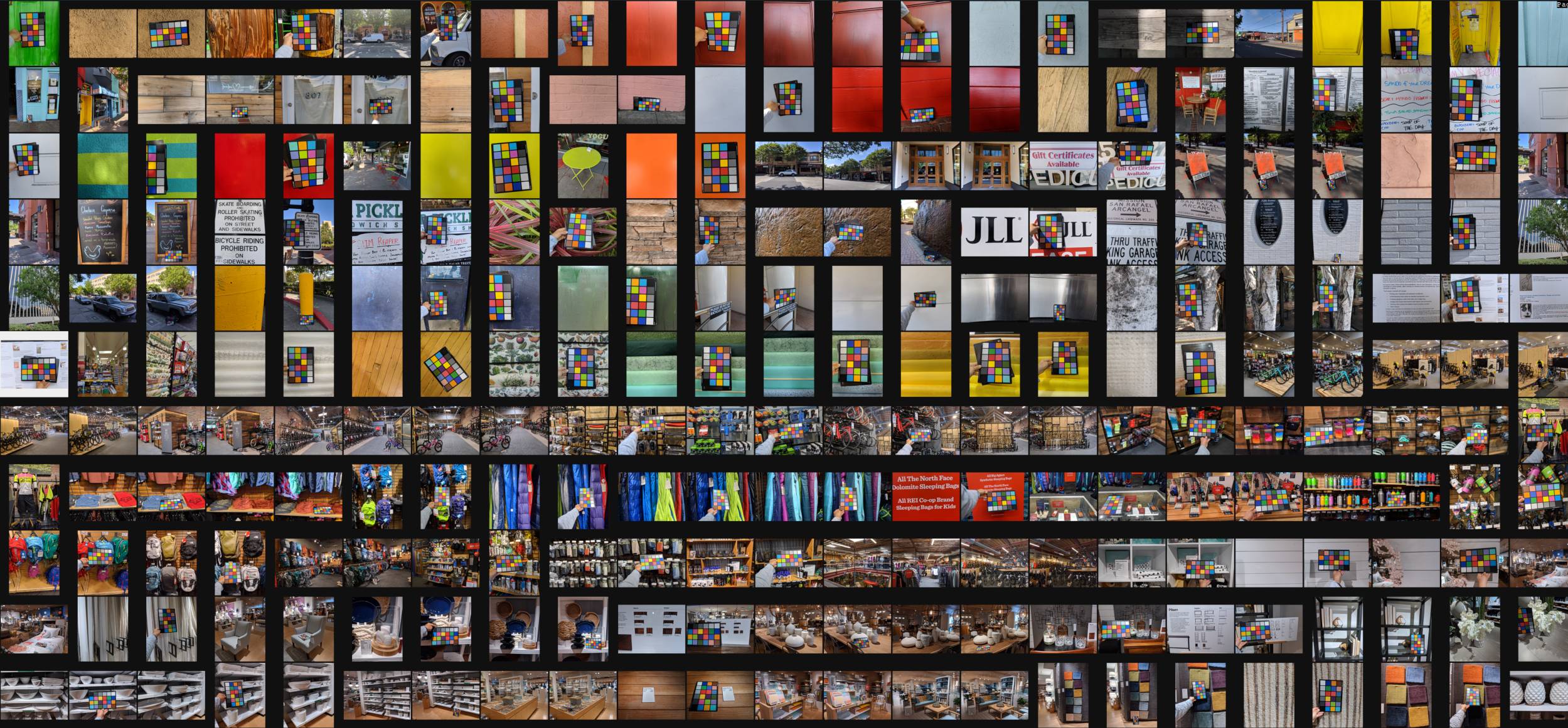}
       \caption{A subset of our training data}
       \label{subfig:awb_tagging_thumbnail} 
    \end{subfigure}
    \begin{subfigure}[b]{\linewidth}
       \includegraphics[width=\linewidth]{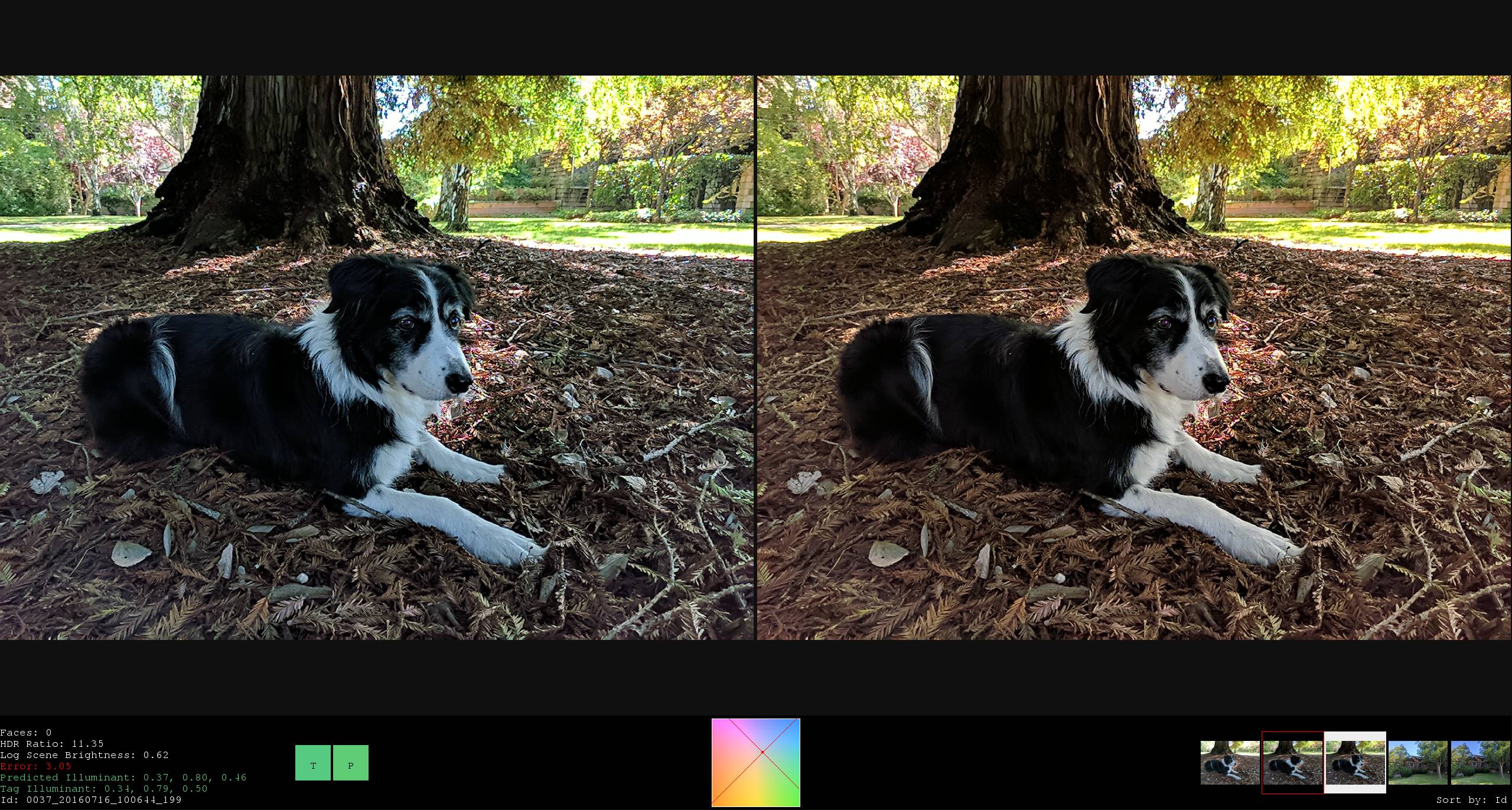}
       \caption{Our AWB tagging tool}
       \label{subfig:awb_tagging_editor}
    \end{subfigure}
    \caption{
    Our learning-based white balance system requires a large and diverse training set, some of which can be seen in (\subref{subfig:awb_tagging_thumbnail}), and a tool for annotating the desired white balance for each image in our dataset, as shown in (\subref{subfig:awb_tagging_editor}).}
    \label{fig:training_set}
\end{figure}

In \fig{fig:training_set} we visualize some of our training data and the tool that was used for annotation.

\subsection{Additional improvements to FFCC}

The "Fast Fourier Color Constancy" (FFCC)~\cite{BarronTsai2017} algorithm that our white balance solution is built upon was modified to improve its speed, quality, and interpretability, as we will describe here.
Instead of providing FFCC with a $2$-channel histogram of pixel and edge intensities for the entire image, we concatenate this histogram with a $2$-channel histogram of pixel and edge intensities for only those pixels that belong to human faces, as predicted by the camera's face detector.
This allows FFCC to explicitly reason about face and non-face content when deciding how to best neutralize the illumination of the image.
\cite{BarronTsai2017} uses a 2-layer neural network to regress from image metadata (exposure time and ISO) to the filter bank that is then used to filter the log-chrominances histograms, which allows the model to indirectly reason about the absolute brightness of the scene and therefore the likely color of the illuminant (i.e.\ when the scene is bright, the illuminant likely has the color of the sun, etc).
We instead directly construct 4-length linearly interpolated histograms from the average log-brightness $L$ of each scene (based on both image content and metadata) and use a linear layer to regress from these histograms to the filter bank.
This allows our model to be interpreted as having 4 separate models (roughly: daylight, indoor, night, and extreme low-light) which are linearly interpolated to produce the model that will be used for a particular scene.
The average log-brightness $L$ is computed as:

\begin{equation} \label{eq:lsb}
    L = \log \left( \underset{i}{\operatorname{median}} \left(\max \left(r^{(i)}, g^{(i)}, b^{(i)} \right) \right) \right) - \log(E)
\end{equation}
where $c^{(i)}$ are RGB values from the input image at channel $c$ and location $i$, and $E$ is the normalized exposure time of the image:
\begin{equation}
    E = \mathrm{exposure\_time} \times \mathrm{analog\_gain} \times \frac{\mathrm{ISO}}{\mathrm{base\_ISO}}
\end{equation}
Where $\mathrm{base\_ISO}$ has been calibrated to be equivalent to ISO 100.
The 4 histogram bin locations for $L$ are precomputed before training as the $12.5$th, $37.5$th, $62.5$th, and $87.5$th percentiles of all $L$ values in our training set, thereby causing our model to allocate one fourth of its capacity to each kind of scene.
The input image we provided to FFCC is a $64\times48$ linear RGB thumbnail, downsampled from the full resolution raw image, and then $3\times3$ median filtered to remove hot pixels.

\subsection{Sensor Calibration}

\begin{figure}
    \centering
    \includegraphics[width=0.8\linewidth]{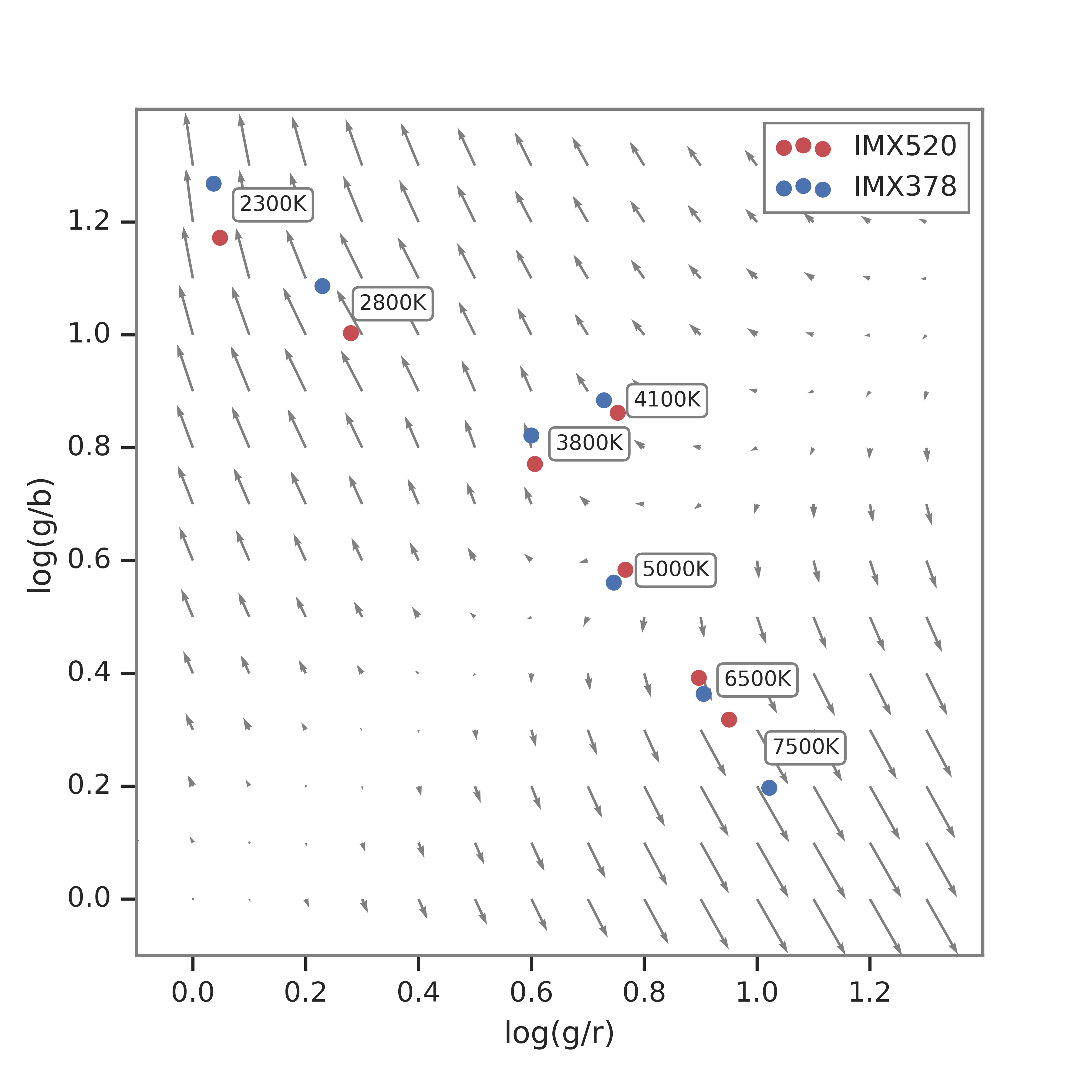}
    \caption{The reference white points from two sensors (IMX520 and IMX378) measured by an X-Rite SpectraLight QC during calibration.
    Our RBF regression estimates a smooth mapping from the IMX520 color space to the IMX378 color space, which is shown here as a quiver plot.
    }
    \label{fig:calib}
\end{figure}

The spectral sensitivity of image sensors varies significantly, both across different generations of sensors and also within a particular camera model due to manufacturer variation (see \fig{fig:calib}).
Designing an AWB algorithm first requires a technique for mapping imagery from any particular camera into a canonical color space, where a single cross-camera AWB model can be learned.

Thankfully, consumer cameras include a calibration step in which the manufacturer records the appearance of $n$ canonical colors as observed by the camera, which we can use to learn a mapping from sensor space to a canonical space. Let $\mathbf{x}$ be a 2-vector that contains $\log(\sfrac{g}{r})$ and $\log(\sfrac{g}{b})$ for some RGB color in the image sensor space, and let $\mathbf{y}$ be a similar 2-vector in the canonical space. We want a smooth and invertible mapping from one space to the other, for which we use radial basis function (RBF) interpolation with a Gaussian kernel.
For each individual camera, given the two $2 \times n$ matrices of observed and canonical log-UV coordinates $\mathrm{X}$ and $\mathrm{Y}$ produced by calibration, we solve the following linear system:
\newcommand{\params}{\Theta}
\begin{gather}
\params = \begin{bmatrix} \varphi\left(\mathrm{X}, \mathrm{X}, \sigma\right) \\ \lambda \mathrm{I} \end{bmatrix} \backslash \begin{bmatrix} \mathrm{Y} - \mathrm{X} \\ \mathrm{0} \end{bmatrix}
\end{gather}
Where $\varphi\left(\mathrm{A}, \mathrm{B}, \sigma\right)$ is an RBF kernel --- a matrix where element $(i,j)$ contains the Gaussian affinity (scaled by $\sigma=0.3$) between the $i$'th column of $\mathrm{A}$ and the $j$'th column of $\mathrm{B}$, normalized per row.
This least-squares fit is regularized to be near zero according to $\lambda=0.1$, which, because interpolation is with respect to relative log-UV coordinates $(\mathrm{Y} - \mathrm{X})$, thereby regularizes the implicit warp on RGB gains to be near $1$. We compute $\params$ once per camera, and when we need to warp an RGB value from sensor space into canonical space, we convert that value into log-UV coordinates and compute
\begin{equation}
  \mathbf{y} = \params^\mathrm{T} \varphi(\mathbf{x}, \mathrm{X}, \sigma)^\mathrm{T} + \mathbf{x}
\end{equation}
This mapping for one camera is shown in Fig. \ref{fig:calib}.
When training our learning-based AWB model, we use this mapping (with $\params$ computed individually for each camera) to warp all image sensor readings into our canonical color space, and learn a single cross-camera FFCC model in that canonical space. When performing AWB on a new image, we use this mapping to warp all RGB values into our canonical color space, use FFCC to estimate white balance gains in that canonical space, and transform the canonical-space white balance gains estimated by FFCC back into the sensor space.
This inverse transformation is performed using Gauss-Newton, which usually converges to within numerical epsilon in 3 or 4 iterations.

\section{Comparisons to other systems}
\label{sec:results_appendix}

\begin{figure}
    \centering
    \includegraphics[width=\linewidth]{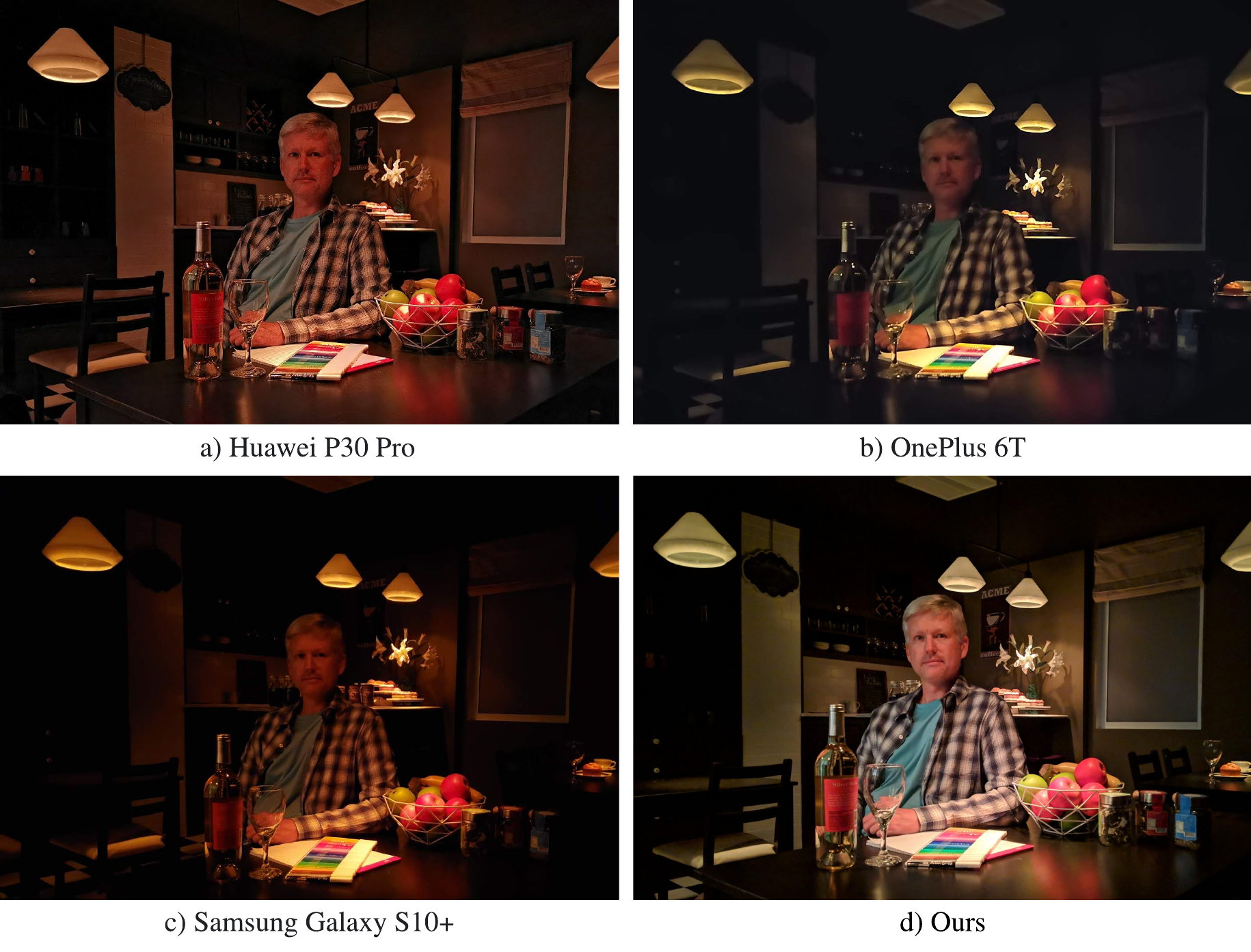}
    \caption{A comparison of our system against the specialized low-light modes of commercial devices. The scene brightness is 0.4 lux and it is illuminated by a dim tungsten light. The devices generally render the skin as too red and do not preserve the details of the face, while some of them do perform well on the other objects in the scene.}
    \label{fig:different_device_comparison_Andy}
    \vspace{0.3in}
    \includegraphics[width=\linewidth]{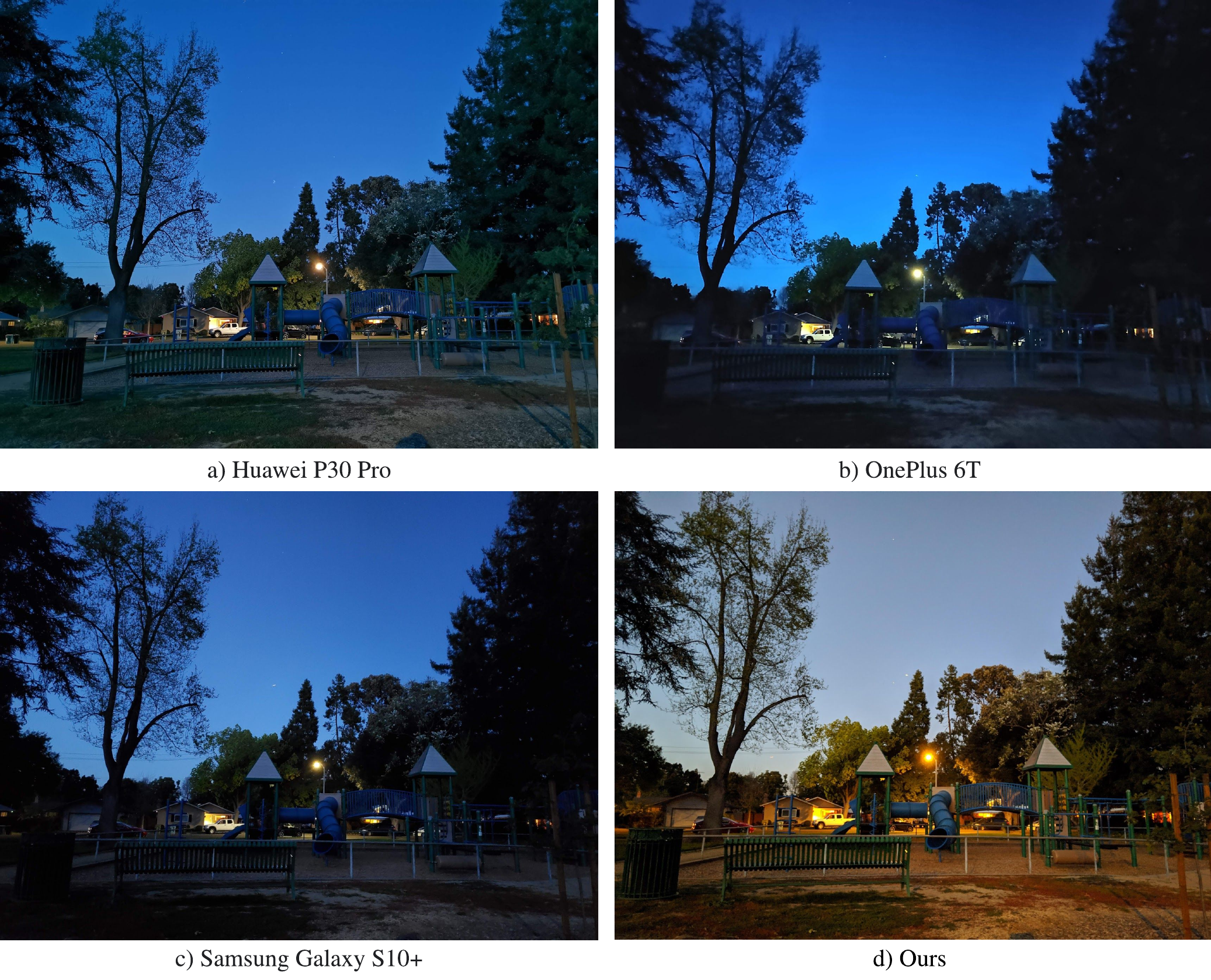}
    \caption{Another comparison of our system against the specialized low-light modes of commercial devices. The scene was captured outdoors shortly after sunset. It is challenging to obtain a natural white balance in this situation. While the other devices render the scene as dark and blue, our processing is able to restore the variety of hues of the scene. These photos were captured with a device which was stabilized by leaning it on a table.}
    \label{fig:different_device_comparison_park}
\end{figure}

In Figures \ref{fig:different_device_comparison_Andy} and \ref{fig:different_device_comparison_park} (as well as in Figure 17 of the main text) we compare photos processed with our on-device pipeline to three commercial devices that have specialized low-light photography modes. These additional examples are provided to demonstrate how image processing pipelines can potentially be inconsistent in their processing. As evidenced by these examples, when the scene is lit by a tungsten light (\fig{fig:different_device_comparison_Andy}) the other devices tend to render the result as too red, while in the evening light (\fig{fig:different_device_comparison_park}), the other devices render the image as too blue. In addition to these differences in colors, there are also significant differences in noise and detail between the different devices in these scenes. 

\begin{figure*}[t!]
\centering
\includegraphics[width=\textwidth]{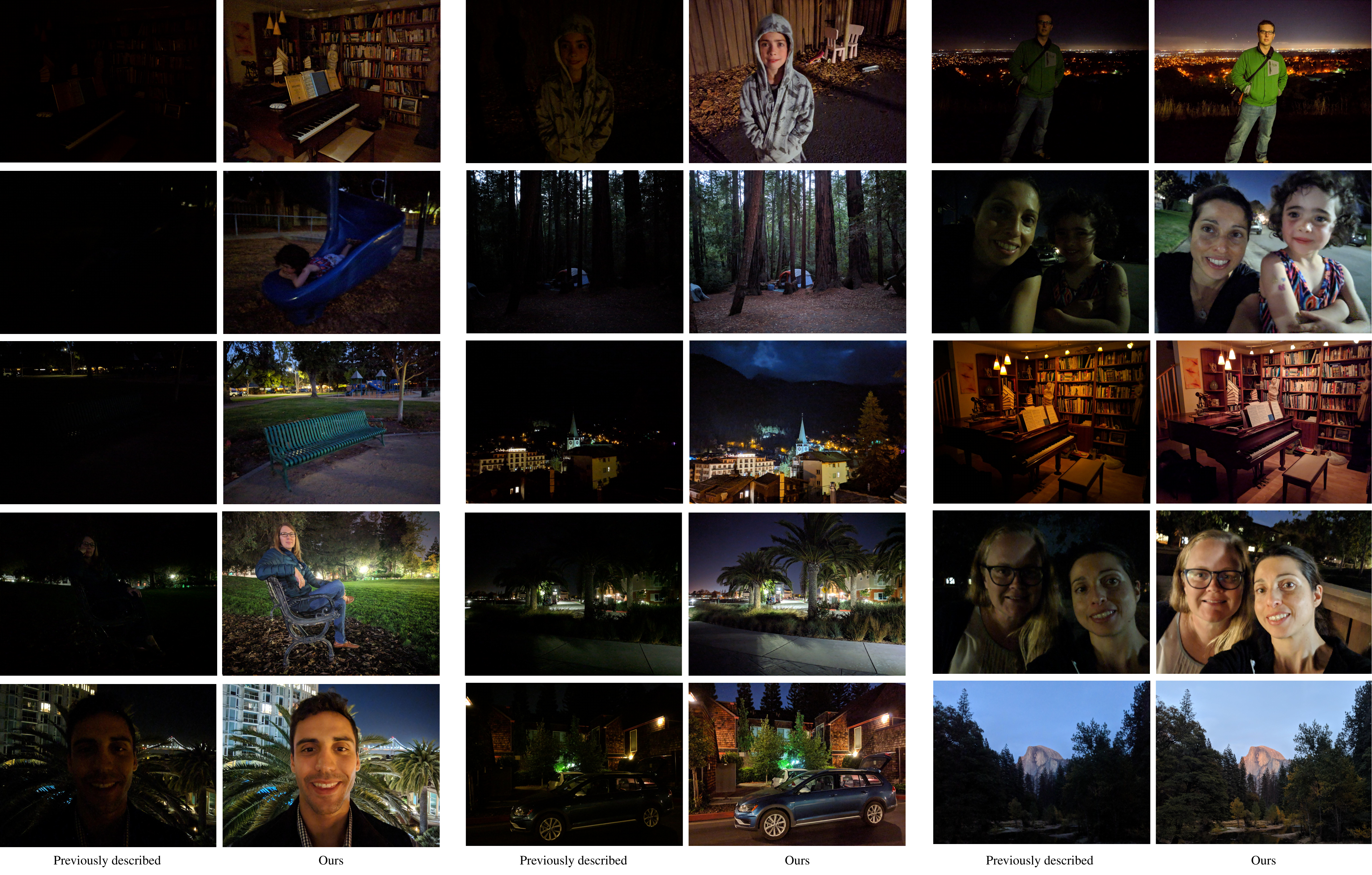}
\caption{A gallery of images comparing the results of the previously described burst processing technique \cite{hasinoff2016burst} and ours across a wide variety of scenes. These photographs were captured handheld, one after the other on the same mobile device, replacing the capture and post-processing modes in between shots, and using either the front or rear facing camera. The images are roughly sorted by scene brightness, the left-most pairs being the darkest and right-most pairs being the brightest. Our approach is able to capture bright and highly detailed photos in very dark environments where the approach of \cite{hasinoff2016burst} breaks down.
The readers are encouraged to zoom in to better see the differences in detail. Simply brightening the images on the left columns would not produce this level of detail (as shown in the main paper) due to the high level of noise. There is also a significant improvement in color composition, owing to our white balance algorithm.}
\label{fig:wall_of_images}
\end{figure*}

Because our system builds upon that of \cite{hasinoff2016burst}, in \fig{fig:wall_of_images} we show a variety of comparisons to that approach on a diverse set of scenes and illuminants. 

\begin{figure*}
    \centering
    \includegraphics[width=\textwidth]{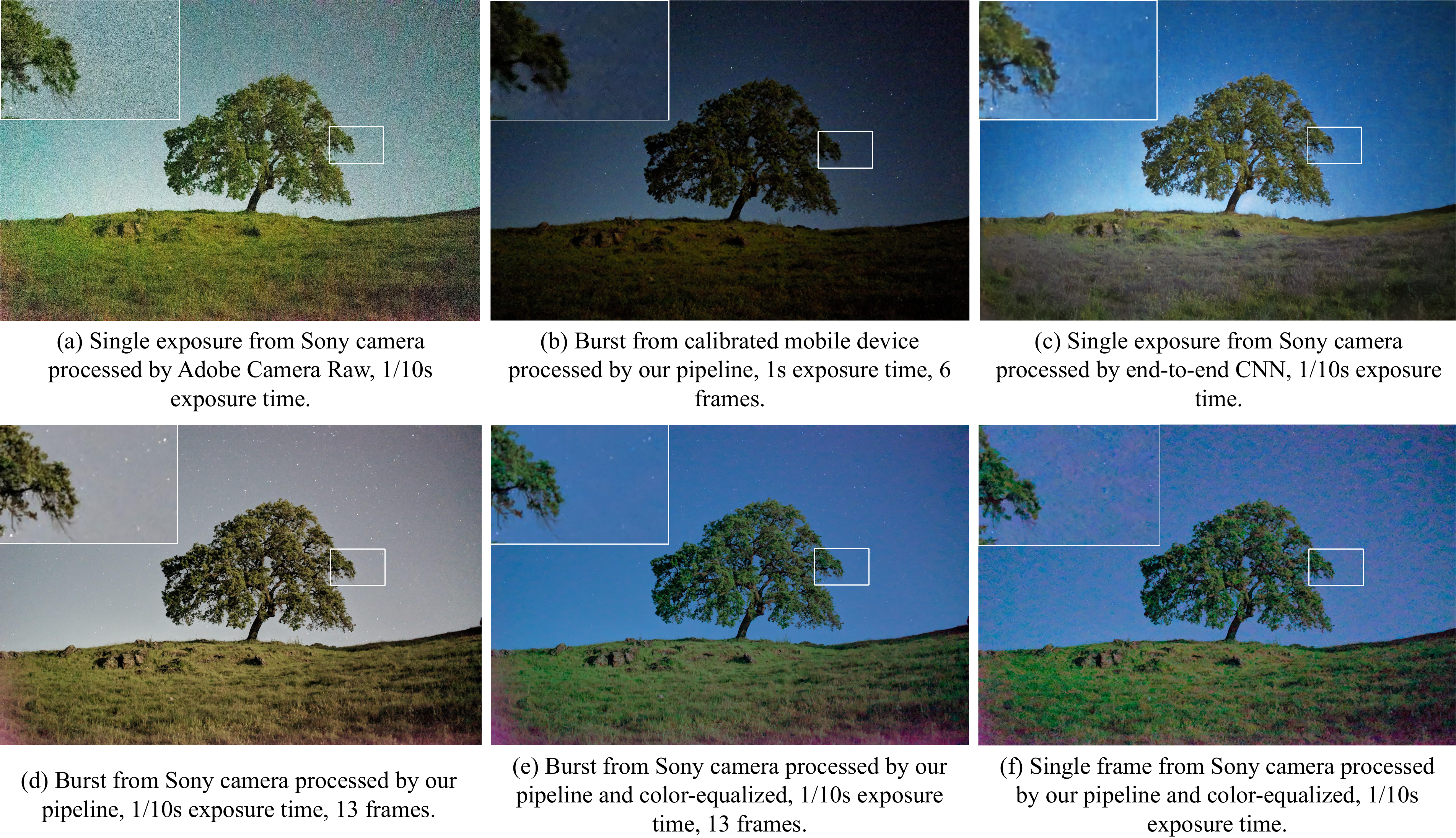}
    \caption{Example of the color matching used for the comparison between the CNN described in \cite{chen2018learning} and our pipeline in the main paper. Raw images were captured with a Sony $\alpha$7S II, the same camera that was used to train the CNN, which was stabilized using a tripod. (a), (b), and (c) demonstrate the varying color rendition from Adobe Camera Raw, our pipeline on frames from a calibrated mobile sensor, and the CNN, respectively. (d) shows the result of our pipeline on 13 raw frames captured with the Sony camera. Because our AWB algorithm is sensor-specific and was not calibrated for this sensor, the resulting colors are different than the result in (d) and closer to that of (a). Because color variation can affect photographic detail perception, the results of our pipeline were color-matched to the results of \cite{chen2018learning} using Photoshop's automatic tool, as shown in (e). The results of our pipeline on a single raw frame (no merging) was also color-matched in (f).}
    \label{fig:results_l2sitd_color_correction}
\end{figure*}

In order to adequately compare the results of our pipeline to the results of the CNN described in \cite{chen2018learning}, we had to apply color matching to our results. Results with different color renditions, including the original rendition of our pipeline, are shown in \fig{fig:results_l2sitd_color_correction}.

\begin{figure*}
    \centering
    \includegraphics[width=\textwidth]{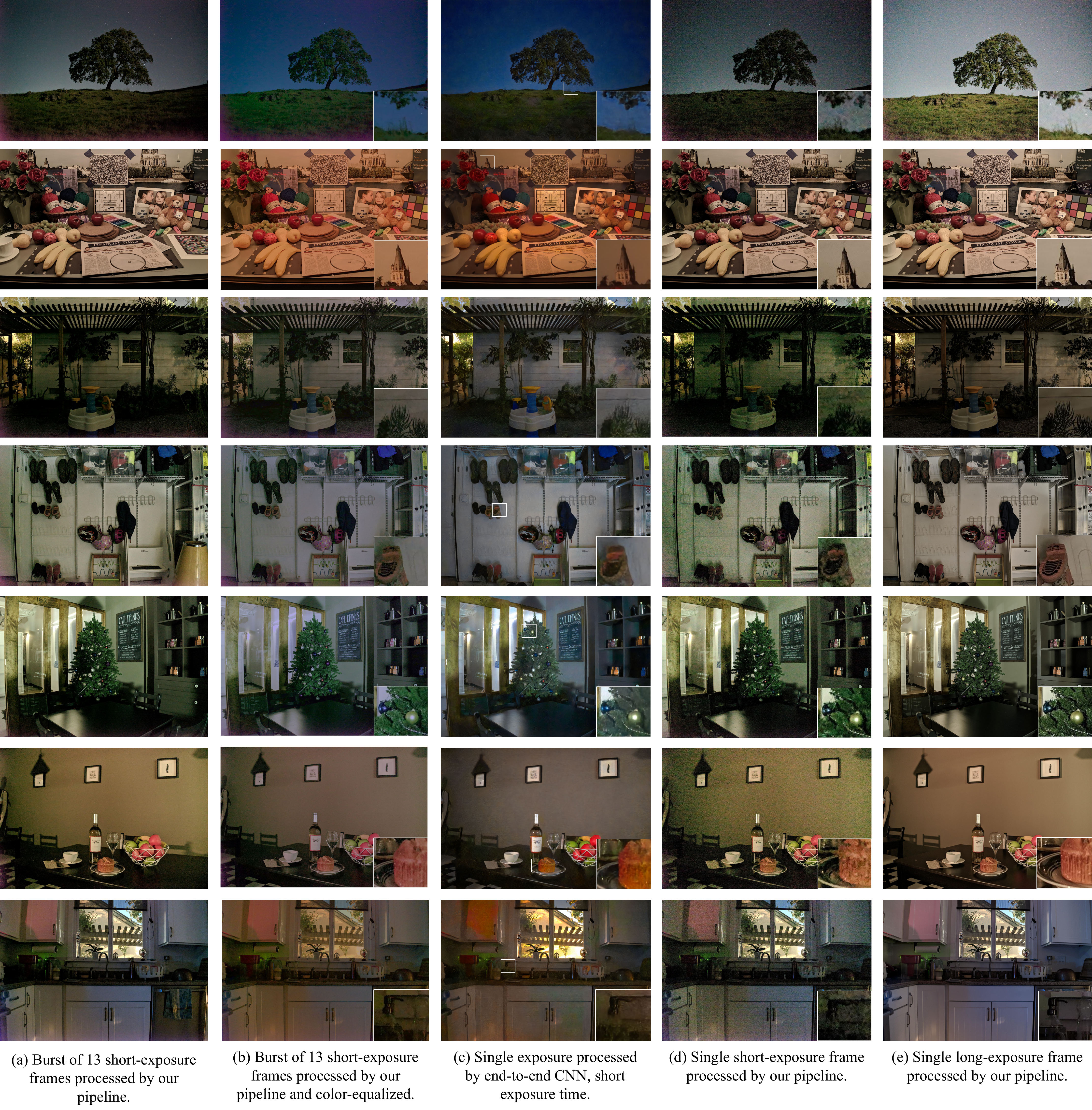}
    \caption{A comparison between the end-to-end trained CNN described in \cite{chen2018learning} and the pipeline in this paper. Raw images were captured with a Sony $\alpha$7S II, which was stabilized using a tripod. The illuminance of these scenes was 0.06-0.5~lux. (a) shows the result of our pipeline on 13 short-exposure frames captured with the Sony camera. Because our AWB algorithm is sensor-specific and was not calibrated for this sensor, the resulting colors are different than the result of the network, which are shown in (c). Because color variation can affect photographic detail perception, we used the color matching described in \fig{fig:results_l2sitd_color_correction} to generate (b). (d) shows the result of processing a single frame through our pipeline (no merging), and (e) serves as a ``ground-truth'' and shows a long exposure frame processed by our pipeline. The short and long exposure times of the frames in the top row are 0.033~s and 0.1~s, respectively, and for the rest of the images they are 0.33~s and 30~s. The advantages of burst processing are evident, as the crops show that the results in our pipeline in (b) produce significantly more detail and less artifacts compared to the CNN output in (c).}
    \label{fig:results_l2sitd_supp}
\end{figure*}

\fig{fig:results_l2sitd_supp} shows additional images that compare our pipeline to an end-to-end trained convolutional neural network (CNN) described in \cite{chen2018learning}. In addition to the result of the CNN and the color-equalized results of our pipeline, we show the result of our pipeline without color-equalization as a reference. We also provide the result of our pipeline on a single long exposure frame, which is provided here as a ``ground-truth'' for the image.

